\documentclass[sigconf, nonacm]{acmart}

\usepackage{balance}

\usepackage[ruled,vlined, linesnumbered]{algorithm2e}
\usepackage{algpseudocode}

\usepackage{tikz}
\usepackage{paralist}
\usepackage{enumerate}
\usepackage[shortlabels]{enumitem}
\usepackage{pifont} 
\newcommand{\cmark}{\ding{51}}%
\newcommand{\xmark}{\ding{55}}
\usepackage{multirow}
\usepackage{lipsum, adjustbox}
\usepackage{xcolor}
\usepackage{colortbl}
\usepackage{tcolorbox}
\definecolor{headergray}{gray}{0.85}
\definecolor{lightgray}{gray}{0.95}
\usepackage{amsmath,scalerel}

\usepackage{tikz}
\usetikzlibrary{arrows.meta,angles,quotes}
\colorlet{linecol}{black!75}
\usepackage{forest}

\tikzset{
  my rounded corners/.append style={rounded corners=2pt},
}

\newcommand*\emptycirc[1][0.7ex]{\tikz\draw (0,0) circle (#1);} 
\newcommand*\halfcircL[1][0.7ex]{%
	\begin{tikzpicture}
		\draw[fill] (0,0)-- (90:#1) arc (90:270:#1) -- cycle ;
		\draw[thick] (0,0) circle (#1);
\end{tikzpicture}}

\newcommand*\fullcirc[1][0.7ex]{\tikz\fill (0,0) circle (#1);} 

\AtBeginDocument{%
  }

\begin{document}

\title{Oops!... They Stole it Again: Attacks on Split Learning}
%


\author{Tanveer Khan}
\orcid{1234-5678-9012}
\affiliation{%
  \institution{Tampere University,}
  \city{Tampere}
  \country{Finland}}
\email{tanveer.khan@tuni.fi}

\author{Antonis Michalas}
\affiliation{%
  \institution{Tampere University,}
  \city{Tampere}
  \country{Finland}}
\email{antonios.michalas@tuni.fi}








\renewcommand{\shortauthors}{Khan et al.}

\begin{abstract}

Split Learning (SL) is a collaborative learning approach that improves privacy by keeping data on the client-side while sharing only the intermediate output with a server. However, the distributed nature of SL introduces new security challenges, necessitating a comprehensive exploration of potential attacks. This paper systematically reviews various attacks on SL, classifying them based on factors such as the attacker's role, the type of privacy risks, when data leaks occur, and where vulnerabilities exist. We also analyze existing defense methods, including cryptographic methods, data modification approaches, distributed techniques, and hybrid solutions. Our findings reveal security gaps, highlighting the effectiveness and limitations of existing defenses. By identifying open challenges and future directions, this work provides valuable information to improve SL privacy issues and guide further research. 
\end{abstract}

\keywords{Privacy-preserving Machine Learning, Split Learning}

\maketitle

\section{Introduction}
\label{sec:intro}


Privacy-preserving Machine Learning (PPML) enables Machine Learning (ML) applications to extract insights without direct access to raw data~\cite{frimpong2024guardml, frimpong2024secrets, khan2021blind, khan2023learning}. Various Privacy-preserving Techniques (PPTs), including cryptographic techniques~\cite{khan2024wildest}, collaborative learning, hybrid models, and data modification approaches, have been explored to enhance privacy. Among these, collaborative learning approaches such as 
Split Learning (SL)~\cite{gupta2018distributed} has gained significant attention for their ability to train ML models across multiple participants without sharing raw data. 

SL splits the model between the client and the server, significantly reducing the computational burden on client devices. 
SL ensures that clients execute the first few layers of the model, while the server handles the more computationally expensive deeper layers. This makes SL an attractive choice for ML-applications on resource-constrained devices. Moreover, since clients and servers do not have access to each other's model portions, SL inherently offers better privacy guarantees. The primary goals of SL are: \begin{inparaenum}[\it (i)] \item preventing raw data from being shared with the server \item maintaining model accuracy comparable to the non-split model \item reducing the computational load on clients by limiting their training to only a few layers \item enabling collaborative learning while keeping each party's model confidential  \end{inparaenum}. 

However, recent studies have shown that there is a privacy leak in SL and is vulnerable to different attacks. Abuadbba \textit{et al.}~\cite{abuadbba2020can} demonstrated that ``intermediate data'' in SL retain significant information about the raw data, which poses a privacy risk. Similarly, Pham \textit{et al.}~\cite{pham2023binarizing} found a similar leakage in Convolutional Neural Network (CNN) based SL, highlighting the potential for Data Reconstruction Attack (DRA). Despite SL's promising advantages, a comprehensive understanding of its security vulnerabilities and potential attack vectors remain unexplored. Although there are existing surveys focusing on Federated Learning (FL), the same comprehensive attention has not been directed towards SL. To bridge this gap, this Systematization of Knowledge (SoK) aims is to systematically analyze the security and privacy risks in SL and to examine various attacks that an adversary $\mathcal{ADV}$ can exploit during training and inference.

\noindent \textbf{Contributions:} The main contributions of this paper are:
\begin{itemize}[left=0pt, nosep]
    \item This SoK provides an extensive review of attacks targeting SL. We categorize these attacks based on multiple dimensions, including who the $\mathcal{ADV}$ is, what type of privacy risk there is, when data leakage occurs, and where the vulnerabilities lie.
    \item \textbf{} We analyze State-of-the-Art (SoTA) defense methods that enhance SL's robustness against privacy threats. This includes cryptographic, data modification, distributed, and hybrid techniques.
\end{itemize}

\subsection{Motivation}
\label{subsec: motivation}

Understanding the security challenges and defenses in SL is essential for building safe and reliable collaborative ML systems. SL allows multiple parties to train ML models together without sharing raw data, helping protect sensitive information. However, SL is vulnerable to various attacks that can expose private data or degrade model performance.

Studying these attacks is important for various reasons. First, it helps identify how privacy can be compromised and how models can be poisoned or manipulated. Second, it supports the design of effective defense strategies that maintain data confidentiality and model integrity. This is especially important in sensitive areas like finance, health, intelligent transport systems and Internet-of-Things (IoTs), where privacy is a top concern. Moreover, as privacy regulations like General Data Protection Regulations (GDPR)~\cite{voigt2017eu} and Health Insurance Portability and Accountability Act (HIPAA)~\cite{edemekong2018health} become stricter, securing SL system is not just a technical need but also a legal one. By exploring both attacks and defenses, researchers can better understand SL's vulnerabilities, develop stronger protections, and support wider adoption. This SoK aims to bring clarity to the current landscape of SL security, helping advance the field and support both academic research and practical deployment.

\subsection{Comparison to Other Surveys}
\label{subsec: comparison}

To clarify our motivation, we provide a summary of related surveys, highlighting their contributions and limitations in relation to PPTs and the collaborative learning. In~\cite{yin2021comprehensive}, the authors present a systematic survey on PPTs in FL, focusing on potential privacy risks and strategies. The authors introduce a 5W scenario-based taxonomy to classify leakage risks and categorize existing privacy-preserving FL (PPFL) methods into four key approaches: encryption, perturbation, anonymization, and hybrid PPFL. The work provides a comprehensive analysis of privacy concerns in FL and offers a structured way to understand existing countermeasures. 
On the other hand,~\cite{matsubara2022split}, presents a detailed review of Split Computing (SC) and Early Exiting (EE), offering comparisons of key approaches and highlight research challenges. The authors summarize different SC and EE methodologies, focusing on model architectures, tasks, and applications, providing a strong foundation for further research in edge-based ML frameworks. Several other studies focus on Vertical FL (VFL). For example,~\cite{khan2022vertical} provides a comprehensive overview of VFL, following its development from basic concepts to real-world applications. It identifies research gaps, suggests improvements, and highlights key areas for future exploration. Similarly,~\cite{liu2024vertical} categorize different VFL settings and PPTs while introducing a unified framework that integrates communication, computation, privacy, and fairness considerations. These works offer valuable insight into privacy and efficiency trade-offs in VFL. 
Yu \textit{et al.}~\cite{yu2024survey} examine privacy threats and countermeasures across different stages of ML process, including data access, pre-processing, training, deployment, and inference. It provides a detailed analysis of privacy attacks and defenses strategies, offering practical guidance for protecting data privacy. 
Jian \textit{et al.}~\cite{jiang2022comprehensive} also provide an in-depth analysis of privacy leakage in VFL, but their focus is limited to the prediction phase. Furthermore, \cite{yu2024survey} discusses privacy threats and countermeasures at different stages of the ML process, including data access, pre-processing, training, deployment, and inference. It provides a detailed analysis of privacy attacks and defense strategies, offering practical guidance to protect data privacy. The work presented in~\cite{hu2025review} provides a systematic review and comparison of the existing SL paradigm, focusing on communication and computation efficiency, label protection and data handling. Similarly,~\cite{thapa2021advancements} explores various aspects of SL and its variants, focusing on optimization performance and communication efficiency. While briefly discusses PPTs like encryption and differential privacy, its main focus is on SL's implementation, real-world applications, and technical challenges.  Another work introduces SLPerf~\cite{zhou2023slperf}, a unified framework and open source library for benchmarking SL methods. It provides a survey of SL paradigms, compare their performance across different datasets, and offer useful insights for future improvements. In addition,~\cite{pham2023data}, offers a valuable contribution by analyzing security threats in SL, examining various attacks that could compromise data privacy, and exploring defense mechanisms to mitigate these risks. While closely related to our work, it remains limited in scope, focusing primarily on Model Inversion Attack (MIA), and Feature-space Hijacking attack (FSHA) while covering only a subset of possible defenses.

Although most prior SoK studies focus primarily on FL, there is a noticeable gap in the literature when it comes to SL. While FL has garnered considerable attention due to its potential in PPML and its 
applicability in decentralized data environments, SL has received comparatively less focus in SoK studies. This disparity can be attributed to the fact that, 
despite the promising paradigm for collaborative learning in situations where data security and privacy are paramount, SL remains less explored in the context of structured surveys. Consequently, existing surveys on SL tend to focus on performance optimization and communication efficiency, rather than systematic 
analysis of security threats or defense mechanisms in SL. This under-representation highlights the need for a more comprehensive analysis of SL, in terms of unique privacy risks, attack vectors, and the tailored countermeasures required to enhance its security and performance. Therefore, our proposed SoK provides a comprehensive examination of SL, including an in-depth analysis of privacy threats and the protective measures employed against them. Specifically, we categorize various attack vectors targeting SL and explore PPTs designed to mitigate these risks. By synthesizing recent advances and identifying key research challenges, this work serves as a valuable reference for ongoing research on privacy-preserving SL (PPSL). It also highlights research gaps and open challenges, aiming to inspire the development of more robust privacy-preserving solutions for real-world applications.

\begin{table*}[h]
    \centering
        \caption{Comparison of our SoK with existing works on Split Learning}
    \label{table: comparisonsoks}
    \renewcommand{\arraystretch}{}  
    \setlength{\tabcolsep}{6pt}  
    \resizebox{\textwidth}{!}{  
    \begin{tabular}{lcccccccccccccc}
        \toprule
        \multicolumn{2}{c}{\textbf{References}} & \multicolumn{5}{c}{\textbf{Collaborative Learning}} & \multicolumn{4}{c}{\textbf{PPTs}} & \multicolumn{4}{c}{\textbf{Attacks}} \\ 
        \cmidrule(lr){3-7} 
        & & \multicolumn{2}{c}{\textbf{Federated Learning}} & \multicolumn{3}{c}{\textbf{Split Learning}} & \multicolumn{4}{c}{\textbf{Techniques}} & \multicolumn{4}{c}{\textbf{SL}} \\ 
        \cmidrule(lr){3-4} \cmidrule(lr){5-7} \cmidrule(lr){8-11} \cmidrule(lr){12-15}
        & & A1 & A2 & B1 & B2 & B3 & C1& C2& C3& C4&D1 &D2 &D3 &D4 \\ 
        \midrule
        Yin \textit{et al.}~\cite{yin2021comprehensive} & & \cmark & \cmark  & \xmark & \xmark & \xmark & \cmark & \cmark & \cmark & \cmark & \xmark & \cmark & \xmark & \xmark\\
        Matsubara \textit{et al.}~\cite{matsubara2022split} & & \xmark & \xmark & \cmark & \xmark & \xmark & \xmark & \xmark & \xmark & \xmark & \xmark & \xmark & \xmark & \xmark \\ 
        Khan \textit{et al.}~\cite{khan2022vertical} & & \xmark &\cmark & \xmark & \xmark & \xmark & \cmark & \cmark & \xmark & \xmark & \xmark & \cmark & \xmark & \xmark \\ 
        Liu \textit{et al.}~\cite{liu2024vertical} & & \xmark & \cmark & \xmark & \xmark & \xmark & \cmark & \cmark & \cmark & \xmark & \cmark & \cmark & \cmark & \xmark \\  
        Ye \textit{et al.}~\cite{ye2024vertical} & & \xmark & \cmark & \xmark & \xmark & \xmark & \cmark & \cmark & \cmark & \xmark & \xmark & \cmark & \cmark & \cmark \\ 
        Yu \textit{et al.}~\cite{yu2024survey} & & \xmark & \cmark & \xmark & \xmark & \xmark & \cmark & \cmark & \cmark & \cmark & \cmark & \cmark & \xmark & \xmark \\ 
        Jian \textit{et al.}~\cite{jiang2022comprehensive} & & \xmark & \cmark & \xmark & \xmark & \xmark & \xmark & \cmark & \cmark & \xmark & \xmark & \xmark & \xmark & \xmark \\ 
        Zhou \textit{et al.}~\cite{zhou2023slperf} & & \xmark & \xmark & \cmark & \cmark & \cmark & \xmark & \xmark & \xmark & \xmark & \xmark & \xmark & \xmark & \xmark \\
         Hu \textit{et al.}~\cite{hu2025review} & & \xmark & \xmark & \cmark & \cmark & \xmark & \xmark & \xmark & \xmark & \xmark & \xmark & \xmark & \xmark & \xmark \\
        Thapa \textit{et al.}~\cite{thapa2021advancements} & & \cmark & \cmark & \cmark & \cmark & \cmark & \textendash & \cmark & \cmark & \cmark & \textendash & \textendash & \textendash & \textendash \\ 
        Pham \textit{et al.}~\cite{pham2023data} & & \xmark & \xmark & \cmark & \cmark & \cmark & \cmark & \cmark & \cmark & \xmark & \cmark & \cmark & \xmark & \xmark \\ 
        Our Work & & \xmark & \cmark & \cmark & \cmark & \cmark & \cmark & \cmark & \cmark & \cmark & \cmark & \cmark & \cmark & \cmark \\ 
        \midrule
        \multicolumn{2}{l}{A1 - Horizontal FL} & \multicolumn{2}{l}{A2 - Vertical FL} \\ \multicolumn{2}{l}{B1 - Vanilla SL} & \multicolumn{2}{l}{B2 - U-shaped SL} & \multicolumn{2}{l}{B3 - Multi-client SL} \\ \multicolumn{2}{l}{C1 - Cryptographic} & \multicolumn{2}{l}{C2 - Data Modification} & \multicolumn{2}{l}{C3 - Distributed} & \multicolumn{4}{l}{C4 - Hybrid} \\ \multicolumn{2}{l}{D1 - Model Inversion} & \multicolumn{2}{l}{D2 -  Inference Attack} & \multicolumn{2}{l}{D3 - Backdoor attacks} & \multicolumn{4}{l}{D4 - Poisoning Attacks} \\ 
        \bottomrule
    \end{tabular}}
\end{table*}

\section{Potential Privacy Leakage Risks in SL}
\label{sec: privacysl}

\begin{figure}
    \centering
    \resizebox{0.49\textwidth}{!}{
\tikzset{every picture/.style={line width=0.75pt}} 

\begin{tikzpicture}[x=0.75pt,y=0.75pt,yscale=-1,xscale=1]

\draw   (209.8,65.6) .. controls (212.89,65.6) and (215.4,68.11) .. (215.4,71.2) -- (215.4,232.4) .. controls (215.4,235.49) and (212.89,238) .. (209.8,238) -- (193,238) .. controls (189.91,238) and (187.4,235.49) .. (187.4,232.4) -- (187.4,71.2) .. controls (187.4,68.11) and (189.91,65.6) .. (193,65.6) -- cycle ;
\draw    (215.2,146.6) -- (223,146.75) ;
\draw   (232.4,52.8) .. controls (232.4,50.81) and (234.01,49.2) .. (236,49.2) -- (343.15,49.2) .. controls (345.14,49.2) and (346.75,50.81) .. (346.75,52.8) -- (346.75,63.6) .. controls (346.75,65.59) and (345.14,67.2) .. (343.15,67.2) -- (236,67.2) .. controls (234.01,67.2) and (232.4,65.59) .. (232.4,63.6) -- cycle ;
\draw   (232.6,103.8) .. controls (232.6,101.81) and (234.21,100.2) .. (236.2,100.2) -- (343.9,100.2) .. controls (345.89,100.2) and (347.5,101.81) .. (347.5,103.8) -- (347.5,114.6) .. controls (347.5,116.59) and (345.89,118.2) .. (343.9,118.2) -- (236.2,118.2) .. controls (234.21,118.2) and (232.6,116.59) .. (232.6,114.6) -- cycle ;
\draw   (231.4,163.4) .. controls (231.4,161.41) and (233.01,159.8) .. (235,159.8) -- (342.9,159.8) .. controls (344.89,159.8) and (346.5,161.41) .. (346.5,163.4) -- (346.5,174.2) .. controls (346.5,176.19) and (344.89,177.8) .. (342.9,177.8) -- (235,177.8) .. controls (233.01,177.8) and (231.4,176.19) .. (231.4,174.2) -- cycle ;
\draw   (232.6,244.4) .. controls (232.6,242.41) and (234.21,240.8) .. (236.2,240.8) -- (344.4,240.8) .. controls (346.39,240.8) and (348,242.41) .. (348,244.4) -- (348,255.2) .. controls (348,257.19) and (346.39,258.8) .. (344.4,258.8) -- (236.2,258.8) .. controls (234.21,258.8) and (232.6,257.19) .. (232.6,255.2) -- cycle ;
\draw    (223.68,58.08) -- (223,249.7) ;
\draw    (223.68,58.08) -- (232.18,58.2) ;
\draw    (346.7,57.93) -- (352.6,58.1) ;
\draw   (360.7,36.7) .. controls (360.7,34.71) and (362.31,33.1) .. (364.3,33.1) -- (469.8,33.1) .. controls (471.79,33.1) and (473.4,34.71) .. (473.4,36.7) -- (473.4,47.5) .. controls (473.4,49.49) and (471.79,51.1) .. (469.8,51.1) -- (364.3,51.1) .. controls (362.31,51.1) and (360.7,49.49) .. (360.7,47.5) -- cycle ;
\draw    (223.68,108.9) -- (232.18,109.02) ;
\draw    (222.86,167.27) -- (231.36,167.4) ;
\draw    (223,249.7) -- (232.75,249.5) ;
\draw   (361.5,65.5) .. controls (361.5,63.51) and (363.11,61.9) .. (365.1,61.9) -- (469.8,61.9) .. controls (471.79,61.9) and (473.4,63.51) .. (473.4,65.5) -- (473.4,76.3) .. controls (473.4,78.29) and (471.79,79.9) .. (469.8,79.9) -- (365.1,79.9) .. controls (363.11,79.9) and (361.5,78.29) .. (361.5,76.3) -- cycle ;
\draw    (352.6,43.1) -- (353.08,70.88) ;
\draw    (352.6,43.1) -- (361.1,43.22) ;
\draw    (353.08,70.88) -- (361.58,71) ;
\draw    (347.7,109.73) -- (353.6,109.9) ;
\draw   (361.7,88.5) .. controls (361.7,86.51) and (363.31,84.9) .. (365.3,84.9) -- (423.6,84.9) .. controls (425.59,84.9) and (427.2,86.51) .. (427.2,88.5) -- (427.2,99.3) .. controls (427.2,101.29) and (425.59,102.9) .. (423.6,102.9) -- (365.3,102.9) .. controls (363.31,102.9) and (361.7,101.29) .. (361.7,99.3) -- cycle ;
\draw   (362.5,117.3) .. controls (362.5,115.31) and (364.11,113.7) .. (366.1,113.7) -- (431.8,113.7) .. controls (433.79,113.7) and (435.4,115.31) .. (435.4,117.3) -- (435.4,128.1) .. controls (435.4,130.09) and (433.79,131.7) .. (431.8,131.7) -- (366.1,131.7) .. controls (364.11,131.7) and (362.5,130.09) .. (362.5,128.1) -- cycle ;
\draw    (353.6,94.9) -- (354.08,122.68) ;
\draw    (353.6,94.9) -- (362.1,95.02) ;
\draw    (354.08,122.68) -- (362.58,122.8) ;
\draw    (346.9,168.52) -- (352.8,168.7) ;
\draw   (363.9,141.3) .. controls (363.9,139.31) and (365.51,137.7) .. (367.5,137.7) -- (452,137.7) .. controls (453.99,137.7) and (455.6,139.31) .. (455.6,141.3) -- (455.6,152.1) .. controls (455.6,154.09) and (453.99,155.7) .. (452,155.7) -- (367.5,155.7) .. controls (365.51,155.7) and (363.9,154.09) .. (363.9,152.1) -- cycle ;
\draw   (364.7,186.1) .. controls (364.7,184.11) and (366.31,182.5) .. (368.3,182.5) -- (451.8,182.5) .. controls (453.79,182.5) and (455.4,184.11) .. (455.4,186.1) -- (455.4,196.9) .. controls (455.4,198.89) and (453.79,200.5) .. (451.8,200.5) -- (368.3,200.5) .. controls (366.31,200.5) and (364.7,198.89) .. (364.7,196.9) -- cycle ;
\draw    (353.2,144.9) -- (353.28,191.48) ;
\draw    (353.2,144.9) -- (363.3,145.02) ;
\draw    (353.28,191.48) -- (364.8,191.5) ;
\draw   (364.7,162.7) .. controls (364.7,160.71) and (366.31,159.1) .. (368.3,159.1) -- (451.8,159.1) .. controls (453.79,159.1) and (455.4,160.71) .. (455.4,162.7) -- (455.4,173.5) .. controls (455.4,175.49) and (453.79,177.1) .. (451.8,177.1) -- (368.3,177.1) .. controls (366.31,177.1) and (364.7,175.49) .. (364.7,173.5) -- cycle ;
\draw    (353.28,168.08) -- (364.8,168.1) ;
\draw    (347.9,249.52) -- (353.8,249.7) ;
\draw   (364.9,212.3) .. controls (364.9,210.31) and (366.51,208.7) .. (368.5,208.7) -- (477.4,208.7) .. controls (479.39,208.7) and (481,210.31) .. (481,212.3) -- (481,223.1) .. controls (481,225.09) and (479.39,226.7) .. (477.4,226.7) -- (368.5,226.7) .. controls (366.51,226.7) and (364.9,225.09) .. (364.9,223.1) -- cycle ;
\draw   (365.7,257.1) .. controls (365.7,255.11) and (367.31,253.5) .. (369.3,253.5) -- (476.26,253.5) .. controls (478.25,253.5) and (479.86,255.11) .. (479.86,257.1) -- (479.86,267.9) .. controls (479.86,269.89) and (478.25,271.5) .. (476.26,271.5) -- (369.3,271.5) .. controls (367.31,271.5) and (365.7,269.89) .. (365.7,267.9) -- cycle ;
\draw    (354.2,215.9) -- (353.71,287.97) ;
\draw    (354.2,215.9) -- (364.3,216.02) ;
\draw    (354.28,262.48) -- (365.8,262.5) ;
\draw   (365.7,233.7) .. controls (365.7,231.71) and (367.31,230.1) .. (369.3,230.1) -- (475.97,230.1) .. controls (477.96,230.1) and (479.57,231.71) .. (479.57,233.7) -- (479.57,244.5) .. controls (479.57,246.49) and (477.96,248.1) .. (475.97,248.1) -- (369.3,248.1) .. controls (367.31,248.1) and (365.7,246.49) .. (365.7,244.5) -- cycle ;
\draw    (354.28,239.08) -- (365.8,239.1) ;
\draw   (365.13,282.59) .. controls (365.13,280.6) and (366.74,278.99) .. (368.73,278.99) -- (475.69,278.99) .. controls (477.67,278.99) and (479.29,280.6) .. (479.29,282.59) -- (479.29,293.39) .. controls (479.29,295.37) and (477.67,296.99) .. (475.69,296.99) -- (368.73,296.99) .. controls (366.74,296.99) and (365.13,295.37) .. (365.13,293.39) -- cycle ;
\draw    (353.71,287.97) -- (365.23,287.99) ;

\draw (210,74) node [anchor=north west][inner sep=0.75pt]  [rotate=-90] [align=left] {{ Privacy Leakage Scenarios}};
\draw (233.2,52) node [anchor=north west][inner sep=0.75pt]   [align=left] {{ \ \ Scenario 1: What}};
\draw (233.4,103) node [anchor=north west][inner sep=0.75pt]   [align=left] {{ \ \ Scenario 2: When}};
\draw (232.2,162.7) node [anchor=north west][inner sep=0.75pt]   [align=left] {{ Scenario 3: Where}};
\draw (236.4,243.8) node [anchor=north west][inner sep=0.75pt]   [align=left] {{ \ \ Scenario 4: Why}};
\draw (364.6,36.4) node [anchor=north west][inner sep=0.75pt]   [align=left] { \ \ Active Attacks};
\draw (365.4,65.2) node [anchor=north west][inner sep=0.75pt]   [align=left] { \ \ Passive Attacks};
\draw (365.6,87.2) node [anchor=north west][inner sep=0.75pt]   [align=left] { \ \ Training};
\draw (366.4,116) node [anchor=north west][inner sep=0.75pt]   [align=left] { \ \ Inference};
\draw (367.8,140) node [anchor=north west][inner sep=0.75pt]   [align=left] { \ \ Intermediate};
\draw (368.6,184.8) node [anchor=north west][inner sep=0.75pt]   [align=left] { \ \ \ \ \ Labels};
\draw (368.6,161.4) node [anchor=north west][inner sep=0.75pt]   [align=left] { \ \ Parameters};
\draw (365.8,211) node [anchor=north west][inner sep=0.75pt]   [align=left] { \ \ Poisoning Attack};
\draw (369.6,255.8) node [anchor=north west][inner sep=0.75pt]   [align=left] { \ \ Model Inversion};
\draw (367.5,232.7) node [anchor=north west][inner sep=0.75pt]   [align=left] { \ \ Backdoor Attack};
\draw (369.03,281.29) node [anchor=north west][inner sep=0.75pt]   [align=left] { \ \ Inference Attack};
\end{tikzpicture}}
    \caption{Taxonomy of Privacy Leakage Scenarios on SL}
    \label{fig:leakage-taxonomy}
\end{figure}
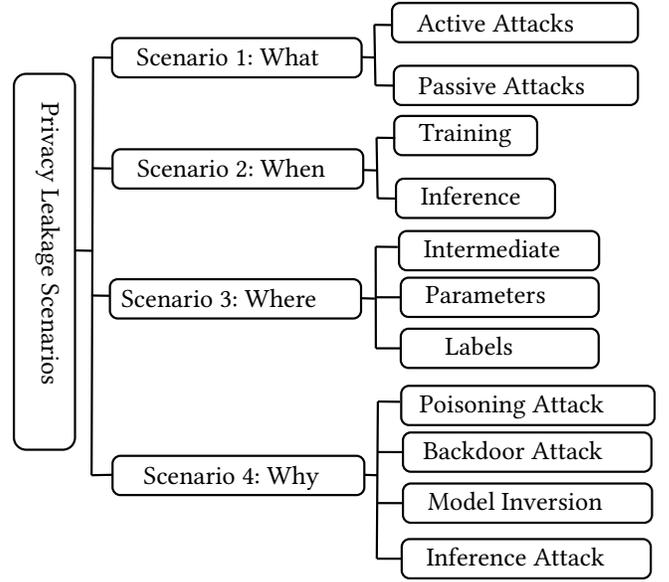

In the following, we provide a taxonomy of privacy leakage scenarios in SL~(\autoref{fig:leakage-taxonomy}). This taxonomy categorizes different attacks and leakage points that may arise during various phases of the SL.

\subsection{What type of privacy attack}
\label{subsec: when}
In SL, clients share intermediate output with the server, raising the fundamental question: Can the server infer private data of clients from the shared intermediate output? To answer this question, researchers have proposed several attack models that assess the privacy risk associated with SL. These attacks can broadly be classified into passive (semi-honest) and active (malicious) attacks.  
  
\noindent \textbf{Passive attacks:} 
It occurs when an $\mathcal{ADV}$ observes the system without modifying it. These attacks try to learn from the shared information without disrupting the training process. In SL, passive attackers monitor the data exchanged between clients and the server, such as intermediate output and gradients.

\noindent \textbf{Active attacks: } 
It involve $\mathcal{ADV}$ who manipulate the learning process to infer private data or degrade model performance. Unlike passive attacks, these attacks actively interfere with the SL process by modifying model parameters, gradients, or training data~\cite{pasquini2021unleashing, gawron2022feature, khowaja2024zeta, fufocusing}. 
In SL, active attacks take two forms: \begin{inparaenum}[\it (i)] \item Malicious clients who manipulate intermediate outputs \item Compromised servers that change model parameters or data labels. \end{inparaenum}

\textit{Since active attacks directly affect model training, they can cause more serious privacy breaches compared to passive attacks. Although passive attacks are harder to detect, active attacks give the attacker more control over the system.}

\subsection{ When might a data privacy leakage occur? }
\label{subsec: when}
In SL, the training phase is divided between a client and a server. The client processes local data through the initial layers of the model and sends the intermediate output to the server. The server computes the final output, calculates the loss, performs backward propagation, and updates the model~\cite{vepakomma2020nopeek}. In the inference phase, unlike training, there is no need for gradient computation or model updates. The client and server simply cooperate to make predictions based on the model that is trained~\cite{liu2024similarity}.

\subsection{Where might a data leakage occur? }
\label{subsec: where}

Privacy is a major concern in SL, as multiple parties collaborate without sharing their raw data. To ensure privacy, both data and model-related information must be protected against malicious threats. These risks arise at different stages of the learning process, such as during the training or inference phase, and affect key components such as input features, output labels, and model architecture, which we explain along with the possible impacts.

\noindent \textbf{Leakage of input features:} 
In SL, although raw data remain with the client, $\mathcal{ADV}$ can attempt to reconstruct the original inputs from the intermediate output shared with the server. This vulnerability makes SL susceptible to DRA, where an $\mathcal{ADV}$ intercepts and analyzes intermediate outputs to infer the original private data~\cite{yang2024uifv,gao2023pcat}. For example, in a medical imaging application, $\mathcal{ADV}$ gaining access to intermediate output could use DRA to approximate the original medical images, leading to unintended exposure of sensitive patient information. Such attacks undermine the fundamental privacy guarantees of SL and highlight the need for robust defense mechanisms~\cite{roth2022split,khowaja2022get}.

\noindent \textbf{Inference from the output labels:} 
In many applications, labels contain highly sensitive information, and their leakage could allow $\mathcal{ADV}$ to deduce private details about the underlying dataset~\cite{fu2022label,kariyappa2021exploit, zhao2024splitaum, liu2024similarity, xie2023label, liu2023distance, erdougan2022unsplit}. For example, in healthcare SL models, if the output labels correspond to a specific diagnosis (e.g., ``cancer detected" or ``no cancer"), an $\mathcal{ADV}$ with access to these labels could infer private medical conditions of patients, even without access to their medical records.  

\noindent \textbf{Model architecture and parameter:} Access to SL's model architecture, parameters, and loss function can lead to attacks such as MIAs~\cite{erdougan2022unsplit, khowaja2022get, li2022ressfl, titcombe2021practical}, poisoning or backdoor attacks, which degrade model performance. 
For example, $\mathcal{ADV}$ 
can modify the loss function to affect its accuracy.

\subsection{Why are the attacks launched?
}
\label{subsec: why}

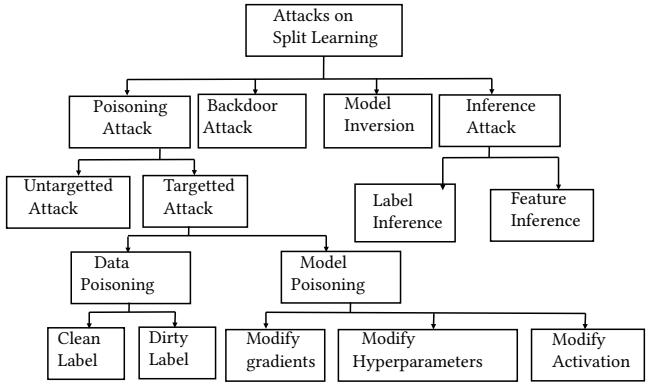
\begin{figure}
    \centering
        \resizebox{0.49\textwidth}{!}{
\tikzset{every picture/.style={line width=0.75pt}} 

\begin{tikzpicture}[x=0.75pt,y=0.75pt,yscale=-1,xscale=1]

\draw   (122.6,91) -- (202.2,91) -- (202.2,125) -- (122.6,125) -- cycle ;
\draw   (171.2,143.33) -- (240,143.33) -- (240,177.33) -- (171.2,177.33) -- cycle ;
\draw   (123.8,193.8) -- (202.4,193.8) -- (202.4,227.8) -- (123.8,227.8) -- cycle ;
\draw   (262.4,193.4) -- (341,193.4) -- (341,227.4) -- (262.4,227.4) -- cycle ;
\draw   (108.4,243.8) -- (158.8,243.8) -- (158.8,277.8) -- (108.4,277.8) -- cycle ;
\draw   (168.6,243.4) -- (219,243.4) -- (219,277.4) -- (168.6,277.4) -- cycle ;
\draw   (225.8,245) -- (291,245) -- (291,279) -- (225.8,279) -- cycle ;
\draw   (300,245) -- (418,245) -- (418,279) -- (300,279) -- cycle ;
\draw    (161.22,183) -- (292,183) ;
\draw    (135.67,233.56) -- (192.44,233.33) ;
\draw    (135.67,233.56) -- (135.67,241.44) ;
\draw [shift={(135.67,244.44)}, rotate = 270] [fill={rgb, 255:red, 0; green, 0; blue, 0 }  ][line width=0.08]  [draw opacity=0] (3.57,-1.72) -- (0,0) -- (3.57,1.72) -- cycle    ;
\draw    (192.44,233.33) -- (192.21,240.89) ;
\draw [shift={(192.11,243.89)}, rotate = 271.81] [fill={rgb, 255:red, 0; green, 0; blue, 0 }  ][line width=0.08]  [draw opacity=0] (3.57,-1.72) -- (0,0) -- (3.57,1.72) -- cycle    ;
\draw    (252.22,233.56) -- (463,234.33) ;
\draw    (252.22,233.56) -- (252.22,241.44) ;
\draw [shift={(252.22,244.44)}, rotate = 270] [fill={rgb, 255:red, 0; green, 0; blue, 0 }  ][line width=0.08]  [draw opacity=0] (3.57,-1.72) -- (0,0) -- (3.57,1.72) -- cycle    ;
\draw    (363,234.33) -- (362.76,241.89) ;
\draw [shift={(362.67,244.89)}, rotate = 271.81] [fill={rgb, 255:red, 0; green, 0; blue, 0 }  ][line width=0.08]  [draw opacity=0] (3.57,-1.72) -- (0,0) -- (3.57,1.72) -- cycle    ;
\draw    (292,183) -- (292,190.89) ;
\draw [shift={(292,193.89)}, rotate = 270] [fill={rgb, 255:red, 0; green, 0; blue, 0 }  ][line width=0.08]  [draw opacity=0] (3.57,-1.72) -- (0,0) -- (3.57,1.72) -- cycle    ;
\draw    (161.22,183.56) -- (161.22,191.44) ;
\draw [shift={(161.22,194.44)}, rotate = 270] [fill={rgb, 255:red, 0; green, 0; blue, 0 }  ][line width=0.08]  [draw opacity=0] (3.57,-1.72) -- (0,0) -- (3.57,1.72) -- cycle    ;
\draw    (128.67,132.93) -- (205,132.69) ;
\draw    (128.67,132.93) -- (128.67,140.5) ;
\draw [shift={(128.67,143.5)}, rotate = 270] [fill={rgb, 255:red, 0; green, 0; blue, 0 }  ][line width=0.08]  [draw opacity=0] (3.57,-1.72) -- (0,0) -- (3.57,1.72) -- cycle    ;
\draw    (182.33,125.5) -- (182.22,132.69) ;
\draw    (201.33,176.83) -- (201.22,183.03) ;
\draw    (308,227.17) -- (307.89,233.36) ;
\draw    (162,227.17) -- (161.89,233.36) ;
\draw    (205,132.69) -- (205,140.26) ;
\draw [shift={(205,143.26)}, rotate = 270] [fill={rgb, 255:red, 0; green, 0; blue, 0 }  ][line width=0.08]  [draw opacity=0] (3.57,-1.72) -- (0,0) -- (3.57,1.72) -- cycle    ;
\draw   (366.63,90.1) -- (446.23,90.1) -- (446.23,124.1) -- (366.63,124.1) -- cycle ;
\draw    (160.67,79.93) -- (401.2,79.29) ;
\draw    (160.67,79.93) -- (160.67,87.5) ;
\draw [shift={(160.67,90.5)}, rotate = 270] [fill={rgb, 255:red, 0; green, 0; blue, 0 }  ][line width=0.08]  [draw opacity=0] (3.57,-1.72) -- (0,0) -- (3.57,1.72) -- cycle    ;
\draw    (401.2,79.29) -- (401.2,86.86) ;
\draw [shift={(401.2,89.86)}, rotate = 270] [fill={rgb, 255:red, 0; green, 0; blue, 0 }  ][line width=0.08]  [draw opacity=0] (3.57,-1.72) -- (0,0) -- (3.57,1.72) -- cycle    ;
\draw   (239.1,30.5) -- (342,30.5) -- (342,64.5) -- (239.1,64.5) -- cycle ;
\draw    (290.33,64.5) -- (290.22,79.69) ;
\draw   (311,149) -- (377,149) -- (377,187) -- (311,187) -- cycle ;
\draw   (400.4,152.6) -- (468.2,152.6) -- (468.2,186.6) -- (400.4,186.6) -- cycle ;
\draw    (368.87,131.53) -- (445.2,131.29) ;
\draw    (399.53,124.1) -- (399.42,131.29) ;
\draw    (368.87,131.53) -- (368.87,150.1) ;
\draw [shift={(368.87,153.1)}, rotate = 270] [fill={rgb, 255:red, 0; green, 0; blue, 0 }  ][line width=0.08]  [draw opacity=0] (3.57,-1.72) -- (0,0) -- (3.57,1.72) -- cycle    ;
\draw    (445.2,131.29) -- (445.2,149.86) ;
\draw [shift={(445.2,152.86)}, rotate = 270] [fill={rgb, 255:red, 0; green, 0; blue, 0 }  ][line width=0.08]  [draw opacity=0] (3.57,-1.72) -- (0,0) -- (3.57,1.72) -- cycle    ;
\draw    (463,234.33) -- (462.76,241.89) ;
\draw [shift={(462.67,244.89)}, rotate = 271.81] [fill={rgb, 255:red, 0; green, 0; blue, 0 }  ][line width=0.08]  [draw opacity=0] (3.57,-1.72) -- (0,0) -- (3.57,1.72) -- cycle    ;
\draw   (427.4,244.8) -- (501.8,244.8) -- (501.8,278.8) -- (427.4,278.8) -- cycle ;
\draw   (290.63,89.7) -- (361.6,89.7) -- (361.6,125) -- (290.63,125) -- cycle ;
\draw    (325.2,79.49) -- (325.2,87.06) ;
\draw [shift={(325.2,90.06)}, rotate = 270] [fill={rgb, 255:red, 0; green, 0; blue, 0 }  ][line width=0.08]  [draw opacity=0] (3.57,-1.72) -- (0,0) -- (3.57,1.72) -- cycle    ;
\draw   (207.63,89.7) -- (278.6,89.7) -- (278.6,125) -- (207.63,125) -- cycle ;
\draw    (245.2,79.49) -- (245.2,87.06) ;
\draw [shift={(245.2,90.06)}, rotate = 270] [fill={rgb, 255:red, 0; green, 0; blue, 0 }  ][line width=0.08]  [draw opacity=0] (3.57,-1.72) -- (0,0) -- (3.57,1.72) -- cycle    ;
\draw   (81.2,144) -- (162.4,144) -- (162.4,178) -- (81.2,178) -- cycle ;

\draw (125,92) node [anchor=north west][inner sep=0.75pt]   [align=left] {\ \ \ \ Poisoning \\ \ \ \ \ \ \ Attack};
\draw (171.2,144) node [anchor=north west][inner sep=0.75pt]   [align=left] {\ \ \ \ Targetted \\ \ \ \ \ Attack};
\draw (128.58,194.8) node [anchor=north west][inner sep=0.75pt]   [align=left] { \ \ \ Data \\Poisoning};
\draw (267.4,194.4) node [anchor=north west][inner sep=0.75pt]   [align=left] { \ \ Model\\Poisoning};
\draw (113.4,244.8) node [anchor=north west][inner sep=0.75pt]   [align=left] {Clean \\Label};
\draw (173.6,244.4) node [anchor=north west][inner sep=0.75pt]   [align=left] { Dirty\\Label};
\draw (225.8,245) node [anchor=north west][inner sep=0.75pt]   [align=left] { \ \ \ \ Modify\\ \ \ \ \ gradients};
\draw (299,245) node [anchor=north west][inner sep=0.75pt]   [align=left] { \ \ \ \ \ Modify \\  \ \ \ Hyperparameters};
\draw (371.03,91.1) node [anchor=north west][inner sep=0.75pt]   [align=left] {\ \ \ \ Inference\\ \ \ \ \ Attack};
\draw (243.5,31.5) node [anchor=north west][inner sep=0.75pt]   [align=left] { \ \ \ \ Attacks on\\ \ \ \ \ Split Learning};
\draw (309.4,153.6) node [anchor=north west][inner sep=0.75pt]   [align=left] { \ \ \ \ Label \\ \ \ \ \ Inference};
\draw (400.4,152.6) node [anchor=north west][inner sep=0.75pt]   [align=left] { \ \ \ \  Feature \\ \ \ \ \ Inference};
\draw (430.6,245.4) node [anchor=north west][inner sep=0.75pt]   [align=left] { \ \ \ Modify \\ \ \ \ Activation};
\draw (290.63,89.7) node [anchor=north west][inner sep=0.75pt]   [align=left] { \ \ \ \  Model \\ \ \ \ \ Inversion };
\draw (200.2,91) node [anchor=north west][inner sep=0.75pt]   [align=left] { \ \ \ \  Backdoor \\  \ \ \ Attack};
\draw (79.2,144) node [anchor=north west][inner sep=0.75pt]   [align=left] {\ \ \ \ Untargetted \\ \ \ \ \ \ Attack};
\end{tikzpicture}
}
    \caption{Taxonomy of Attacks on Split Learning}
    \label{fig:attack-taxonomy}
\end{figure}

$\mathcal{ADV}$ exploit vulnerabilities in SL to achieve malicious objectives such as extracting private information, degrading model performance, or embedding backdoor for targeted manipulation. These attacks (see~\autoref{fig:attack-taxonomy}) can be categorized based on their execution strategy, either from the client or server side -- and include inference attacks, which aim to reconstruct sensitive data; poisoning attacks, which manipulate training data or model updates; and backdoor attacks, where $\mathcal{ADV}$ implement hidden triggers to control model behavior. A particularly concerning attack is MIA, where an $\mathcal{ADV}$ reconstructs private training data by exploiting model output or intermediate output, posing a significant threat to SL's privacy and security. 
Regardless of their origin, such attacks pose serious threats to the privacy and security of SL frameworks.

\subsubsection{Poisoning attack (PA)}
\label{subsubsec: poisoning}
PA in ML are a significant threat that aims to manipulate a model’s training process toward malicious ends~\cite{tian2022comprehensive}. These attacks involve injecting corrupted data points, targeting specific classes to reduce the accuracy of certain predictions, or altering the model’s decision boundaries. 
Poisoning attacks in SL can be classified into two classes:

\noindent \textbf{Targeted poisoning attacks (TPA):} The $\mathcal{ADV}$ aims to manipulate the model to produce incorrect predictions for a specific class by replacing its label with a target class label. This reduces the accuracy of the classifier for the target source class, without affecting others. In the SL setting, an $\mathcal{ADV}$ can manipulate the activation data sent from the client to the sever or inject poisoned samples into the local dataset. Since the server does not have access to raw data, detecting these attacks is more challenging, leading to consistent misclassifications on the server-side. 
As can be seen in~\autoref{fig:attack-taxonomy}, TPAs~\cite{ismail2023analyzing} are further classified into:

\textbf{Data poisoning attack (DPA):} An $\mathcal{ADV}$ manipulate training data to degrade model performance or induce misclassifications. In SL, this attack is particularly effective when client control their local datasets, as they can induce poisoned samples undetected. DPA broadly take the form of \begin{inparaenum}[\it (i)] \item \textbf{Clean Label Data Poisoning (CLDP):} $\mathcal{ADV}$ insert subtly altered but correctly labeled inputs, which mislead the model into learning incorrect representations~\cite{gajbhiye2022data}. This is particularly dangerous in SL, where the server never sees raw data, making detection difficult, and \item \textbf{Dirty Label Data Poisoning (DLDP):} $\mathcal{ADV}$ injects mislabeled data into the training set. Since SL relies on distributed training~\cite{rossolini2024edge, rossolini2025exploiting}, the server cannot independently verify the correctness of the labels, increasing the vulnerability. 
\end{inparaenum}

\textbf{Model Poisoning Attack (MPA):} An $\mathcal{ADV}$ directly manipulates model parameters to cause undesirable behavior~\cite{kumaar2024fortifying, khan2022security}. In SL, $\mathcal{ADV}$ can manipulate intermediate outputs, gradients, or hyper-parameters exchanged between client and server. It can disrupt the model, causing biased predictions, or lower accuracy.
To demonstrate such threats, He \textit{et al,}~\cite{he2024advusl} introduce AdvUSL, a targeted attack U-shaped SL that trains a lightweight surrogate model using limited labeled data, learns to match the intermediate output with their labels, and then uses this information to launch targeted attacks on training process.

\noindent \textbf{Untargeted poisoning attacks (UPA):} 
Unlike TPAs, UPAs randomly flip a set of labels or change all labels to a single random class, significantly reducing the classifier accuracy. In SL, this attack can occur on the client- or server-side by introducing malicious gradient updates,  causing incorrect gradients to propagate and reducing the overall model accuracy. A study by Gajbhiye \textit{et al.}~\cite{gajbhiye2022data}, found that poisoning attacks are a major threat to SL, with UPA causing more damage to model accuracy than TPAs. Similarly, Fan \textit{et al,} propose SLADV~\cite{fan2023robustness}, a non-targeted attacks in which an $\mathcal{ADV}$ -- typically the server -- modifies model updates, leading to incorrect predictions and weakening the system's security. 

\subsubsection{Backdoor attacks}
\label{subsubsec: backdoor}

Poisoning-based attacks are the dominant method of creating backdoor attacks, where an $\mathcal{ADV}$ aims to implant a specific pattern or trigger in the model. The idea is that when this trigger appears on the input during testing, the model outputs a target class while behaving normally for other inputs. Typically, this is done by poisoning the training data, where $\mathcal{ADV}$ alters some of the data samples with a trigger pattern, causing it to misclassify the poisoned inputs during the testing~\cite{tajalli2023feasibility}.

In SL, there are two main scenarios for backdoor attacks. The first involves a \textit{client-side backdoor attack}, where an $\mathcal{ADV}$ is one of the clients or has access to the client's data. In this case, an $\mathcal{ADV}$ can manipulate the local dataset by inserting poisoned samples with a specific trigger and a target label. For example, in the U-shaped SL protocol, an $\mathcal{ADV}$ can easily alter the training data by changing a clean sample to a target class. However, in the vanilla SL protocol, where the features and labels are separated between the client and the server, constructing a backdoor sample is more challenging~\cite{yu2023backdoor}. $\mathcal{ADV}$ needs to infer the label information and then manipulate the client's data to align with the target class. He \textit{et al.}~\cite{he2023backdoor} demonstrated a client-side attack that injects a backdoor through trigger vectors without altering labels. Bai \textit{et al.}~\cite{bai2023villain} proposed VILLIAN, a backdoor attack framework for vanilla SL that enables malicious clients to inject a backdoor without knowing the data labels. Yu \textit{et al.}~\cite{yu2024chronic} designed an attack using a shadow model to transfer backdoor capabilities to the client, proposing the SFI framework (Steal, Finetune and Implant) to execute stealthy backdoor attacks with minimal assumptions. 

The second scenario is a \textit{server-side backdoor attack}, where all clients are honest. The server might inject a backdoor into the model by subtly manipulating the training process without direct access to the client data. Because the server cannot directly access the training data in SL, this makes it more challenging for an $\mathcal{ADV}$ to execute a backdoor attack. However, there have been attempts to bypass these limitations by training surrogate models or using autoencoders to inject a backdoor into the server-side model~\cite{tajalli2023feasibility}. Yu \textit{et al.}~\cite{yu2023backdoor} showed SL's vulnerability to backdoor attacks by using a shadow model to manipulate gradients. A more effective approach is the Stealthy Backdoor Attack (SBAT)~\cite{pu2024stealthy}, where the attacker injects a trigger embedded directly into the server network, avoiding any modification to gradients or training data. This method ensures both stealth and effectiveness by executing malicious operations after training, making it difficult to detect. Fan \textit{et al.}~\cite{fan2023robustness} introduced a two-stage attack using a surrogate model and adversarial examples, revealing SL's vulnerability to server-side backdoor attack. Advancing this area further, Pu \textit{et al,}~\cite{pu2024dullahan} develop Dullahan, a backdoor attack that embeds trigger and taget embeddings within the server network, achieving high attack success without degrading model utility.

\paragraph{\textbf{Discussion}} These attacks in SL present significant security risk. PA aim to degrade model performance or manipulate decision-making, while backdoor attacks introduce hidden vulnerabilities that can be exploited later. Unlike traditional centralized learning, SL's architecture makes 
such attacks harder to detect, as raw data is not shared, and only intermediate output 
is exchanged.  

\subsubsection{Model inversion and Data reconstruction attack}
\label{subsubsec: modelinversion}    

MIA and DRA pose a significant threat to privacy in SL by allowing $\mathcal{ADV}$ to reconstruct sensitive client data from shared feature representations. In SL, the client processes raw data and sends intermediate output to the server, which then completes the training process. However, these shared outputs can be exploited to infer the original raw data, making SL vulnerable to MIA~\cite{erdougan2022unsplit}.
MIA typically works by inverting the model's output or intermediate output to reconstruct details about the raw data. Although early MIAs focus on the model's output, recent research has shown that attacks on hidden layers are more effective, as these representations retain structural similarities to the original inputs. Several techniques have been developed to exploit this vulnerability, each refine the process of recovering private data from SL models. 
For example, query-free methods train shadow  models on similar data to imitate target behavior and recover private inputs, even without access to gradient or model queries. This highlight  the persistent risk of data leakage  in collaborative learning frameworks regardless of $\mathcal{ADV}$ capabilities. FSHA~\cite{pasquini2021unleashing,gawron2022feature}, demonstrates that a malicious server can hijack the client's feature extraction process. The attacker trains an autoencoder on a shadow dataset to learn mappings between feature space and input space. During SL training, the attacker gradually forces the feature representation of the client model to align with those of a shadow dataset, allowing the server to reconstruct client data. However, FSHA is intrusive and often disrupts the SL process, making it detectable by the client \cite{erdogan2022splitguard}. More recent work enhances this attack by improving reconstruction quality, adapting it to vertically partitioned data, and introducing flexible training modes~\cite{yu2024sia}. A more stealthy approach is SplitSpy~\cite{fufocusing}, where an $\mathcal{ADV}$ maintains a legitimate SL model that operates normally while secretly detecting and manipulating certain feature representations. By filtering out manipulated gradients and sending deceivable gradients back to the client, $\mathcal{ADV}$ gradually guides the client model to a state where its feature representations can be inverted more easily, making DRA more efficient.
ZETA~\cite{khowaja2024zeta} introduces a shadow model trained on public data to mimic client's data. A Generative Adversarial Network (GAN) is employed between shadow model and server, refining the alignment between the client's and the shadow model's output. This allows $\mathcal{ADV}$ to generate reconstructions of private client data without disrupting SL training, making detection difficult. Other GAN-based techniques~\cite{yang2024uifv} use pre-trained GANs to explore hidden patterns and find examples that closely resemble the actual input.

A more advanced attack, Pseudo-Client Attack (PCAT)~\cite{gao2023pcat}, does not require knowing the client's model architecture. Instead, it improves its understanding of the client's data by optimizing the server-side model to reconstruct private inputs, making it effective even with limited information. Similarly, Feature-Oriented Reconstruction Attack (FORA)~\cite{xu2024stealthy}, also does not require access to client's model. Instead, it focuses on learning the unique features of the client's data by training a substitute model that mimics these features. This allows $\mathcal{ADV}$ to reconstruct the original data from the intermediate output, making it effective even in complex SL scenarios. 
More recently, the Unified InverNet Framework in VFL~\cite{yang2024uifv}, introduced an alternative approach that removes the dependency on gradient or model information. Instead, UIFV directly reconstructs private data from intermediate feature vectors using an inverse network (InverNet), making it highly adaptable to VFL settings where different parties hold distinct features. 

\textbf{Discussion:} The evolution of MIA in SL has shifted from simplistic reconstructions based on output predictions to sophisticated inversion techniques targeting hidden-feature representations. Early attacks like FSHA were easily detectable due to their impact on SL training, whereas modern methods like PCAT, FORA, and GAN-based approaches operate with minimal disruption, making them significantly harder to defend against. As SL continues to be adopted for PPML, it is important to understand and mitigate MIA risks.

\subsubsection{Inference attacks (IA)}
\label{subsubsec: inference-attack}
IA in SL pose serious privacy concerns by allowing an $\mathcal{ADV}$ to extract private information from intermediate output exchanged between clients and the server. These attacks can be categorized into the following.

\noindent \textbf{Label Inference Attack (LIA):} In U-shaped SL, clients do not directly share labels, but the gradients received during training contain hidden-label information. Since loss gradients depend on the difference between the predicted and actual labels, $\mathcal{ADV}$ can analyze these gradients to infer private labels. An approach to LIA is the model completion, where an $\mathcal{ADV}$ reconstructs a complete model using auxiliary labeled data~\cite{fu2022label}. By fine-tuning this model, they can improve the accuracy of label inference. More advanced attacks manipulate the training process by modifying gradients or optimizing attack-specific loss functions, forcing the SL model to rely more heavily on the attacker's model. Another LIA method, Exploit~\cite{kariyappa2021exploit}, replaces private labels with a surrogate model and labels, using gradient-based optimization to learn the true labels over time. Similarly, distance-based attacks~\cite{liu2023distance} measure gradient similarity to infer labels and the clustering-based method improves efficiency~\cite{liu2024similarity}. 
For regression tasks, where labels are continuous, a novel learning-based attack~\cite{xie2023label} integrates gradient analysis with additional regularization to accurately infer labels. To enhance scalability, SplitAUM~\cite{zhao2024splitaum} introduces an auxiliary model that learns from intermediate features and uses semi-supervised clustering to infer labels without direct access to true labels.

\noindent \textbf{Feature Inference Attack (FIA):} In FIA, the goal of $\mathcal{ADV}$ is to reconstruct the client's input features using shared gradients or intermediate output. The $\mathcal{ADV}$ exploits the fact that deep learning models preserve information about input features even in intermediate layers. Simulator Decoding with Adversarial Regularization (SDAR)~\cite{zhu2023passive}, trains a simulator model to imitate the client's feature extraction process, using adversarial regularization to refine the reconstruction. Another technique, Feature Sniffer~\cite{luo2023feature}, transfers identity information to an auxiliary dataset, allowing the attacker to stealthily extract private data. Exact~\cite{qiu2023exact} assumes that the client features belong to discrete categories and builds a list of possible combinations of feature-label. It uses activation maps and gradient patterns to reconstruct the most likely features. 

\paragraph{\textbf{Discussion}} IA in SL reveal critical privacy vulnerabilities, demonstrating that even when raw data remains on the client-side, attackers can still extract labels and features from exchanged information. LIA exploits gradient information to reconstruct private labels, while FIA reconstruct input features from intermediate activations.

\noindent \textbf{Analysis:} Poisoning and backdoor attacks in SL pose a significant security risk but differ in their approach and impact. Poisoning attacks aim to degrade model performance or manipulate decision making, while backdoor attacks introduce hidden vulnerabilities that can be exploited later. The SL architecture, where only intermediate outputs are exchanged, and raw data remains private, makes these attacks harder to detect. This complexity in detection is similar to the challenges posed by MIA, which have evolved significantly in SL.  
MIA in SL has shifted from simpler attacks, such as FSHA, which were easily detectable, to more sophisticated attacks, such as PCAT, FORA, and GAN-based approaches. These advanced MIA cause minimal disruption to SL, making them significantly harder to defend against. Additionally, IAs, in SL reveal critical privacy vulnerabilities in SL. Even when raw data remain on the client-side, $\mathcal{ADV}$ can still extract sensitive information from the exchanged intermediate output.

\begin{table*}[!]
\caption{\fullcirc~means the attack needs the corresponding assumption, \emptycirc~means it does not, and \halfcircL~means it is not required but helpful. "Data assumption" refers to access to an auxiliary dataset of same distribution, and "Model assumption" refers to knowing the client model's architecture or weights. \cmark~denotes use of the technique, \xmark~means it does not, and \textendash~means it is unspecified.
}
\label{table: attacks comparison}
\resizebox{0.97\textwidth}{!}
{
\begin{tabular}{lllllllllllllllllllllllllllllll}
\hline
\multirow{2}{*}{\large{Attacks}} &  & \multirow{2}{*}{Type} &  & \multirow{2}{*}{Name} &  & \multicolumn{2}{l}{PA} &  & \multicolumn{4}{l}{SL} &  & \multicolumn{2}{l}{MA} &  &  & \multicolumn{2}{l}{DA} &  & \multicolumn{2}{l}{MS} &  & \multicolumn{2}{l}{GL} &  & \multicolumn{2}{l}{PS} &  & \multirow{2}{*}{Approach} \\ \cline{7-8} \cline{10-13} \cline{15-16} \cline{19-20} \cline{22-23} \cline{25-26} \cline{28-29}
 &  &  &  &  &  & PA1 & PA2 &  & VSL & USL & MCSL & FL-SL &  & MA1 & MA2 &  &  & DA1 & DA2 &  & MS1 & MS2 &  & GL1 & GL2 &  & PS1 & PS2 &  &  \\ \hline
\multirow{6}{*}{PA} &  & \multirow{3}{*}{DPA} &  &Ismail \textit{et al.}~\cite{ismail2023analyzing} &  & \cmark  & \xmark &  & \xmark & \xmark & \xmark & \cmark &  & \textendash & \textendash &  &  & \textendash & \textendash &  & \cmark & \xmark &  & \textendash & \textendash &  & \xmark & \cmark &  & \begin{tabular}[c]{@{}l@{}}Poison the \\ training data\end{tabular} \\
 &  &  &  &He \textit{et al.}~\cite{he2023backdoor} &  & \cmark & \xmark &  & \cmark & \xmark & \xmark & \xmark &  & \textendash & \textendash &  &  & \fullcirc & \fullcirc &  & \cmark & \xmark &  & \textendash & \textendash &  & \xmark & \cmark &  & \begin{tabular}[c]{@{}l@{}}Replace the local embedding \\ with a trigger vector\end{tabular} \\ \cline{2-31} 
 &  & \multirow{3}{*}{MPA} &  &kumaar \textit{et al.}~\cite{kumaar2024fortifying} &  & \cmark & \xmark &  & \xmark & \xmark & \xmark & \cmark &  & \textendash & \textendash &  &  & \textendash & \textendash &  & \cmark & \xmark &  & \textendash & \textendash &  & \xmark & \cmark &  & \begin{tabular}[c]{@{}l@{}}Distance-based\\  attack strategies\end{tabular} \\
 &  &  &  &Khan \textit{et al.}~\cite{khan2022security} &  & \cmark & \xmark &  & \xmark & \xmark & \xmark & \cmark &  & \textendash & \textendash &  &  & \textendash & \textendash &  & \cmark & \xmark &  & \textendash & \textendash &  & \xmark & \cmark &  & \begin{tabular}[c]{@{}l@{}}Targeted \\ adversarial attack\end{tabular} \\
 &  &  &  &AdvUSL~\cite{he2024advusl} &  & \cmark & \xmark &  & \xmark & \cmark & \xmark & \xmark &  & \fullcirc & \emptycirc &  &  & \fullcirc & \fullcirc &  & \xmark & \cmark &  & \xmark & \cmark &  & \xmark & \cmark &  & \begin{tabular}[c]{@{}l@{}}Optimization based \\ model poisoning attack\end{tabular} \\
 &  &  &  &MISA~\cite{wan2024misa} &  & \cmark & \xmark &  & \xmark & \xmark & \xmark & \cmark &  & \fullcirc & \emptycirc &  &  & \fullcirc & \emptycirc &  & \cmark & \cmark &  & \xmark & \cmark &  & \xmark & \cmark &  & \begin{tabular}[c]{@{}l@{}}Poison both \\ model\end{tabular} \\ \hline
\multirow{8}{*}{Backdoor} &  & \multirow{3}{*}{\begin{tabular}[c]{@{}l@{}}Client\\  side\end{tabular}} &  &Yu \textit{et al.}~\cite{yu2023backdoor} &  & \cmark & \xmark &  & \cmark & \cmark & \textendash & \textendash &  & \emptycirc & \emptycirc &  &  & \emptycirc & \emptycirc &  & \cmark & \cmark &  & \textendash & \textendash &  & \xmark & \cmark &  & \begin{tabular}[c]{@{}l@{}}Insert backdoor samples\\into training data, Manipulate\\client model optimization\\to encode backdoor\end{tabular} \\
 &  &  &  & VILLIAN\cite{bai2023villain} &  & \cmark & \xmark &  & \cmark & \xmark & \xmark & \xmark &  & \emptycirc & \emptycirc &  &  & \fullcirc & \emptycirc &  & \cmark & \xmark &  & \textendash & \textendash &  & \xmark & \cmark &  & \begin{tabular}[c]{@{}l@{}}Injects a backdoor \\ into the server model\end{tabular} \\ \cline{3-31} 
 &  & \multirow{5}{*}{\begin{tabular}[c]{@{}l@{}}Server\\ side\end{tabular}} &  &Tajalli \textit{et al.}~\cite{tajalli2023feasibility} &  & \cmark & \xmark &  & \cmark & \xmark & \cmark & \xmark &  & \fullcirc & \textendash & &  & \textendash & \fullcirc &  & \xmark & \cmark &  & \textendash & \textendash &  & \xmark & \cmark &  & \begin{tabular}[c]{@{}l@{}}Using surrogate client\\  and auto encoder\end{tabular} \\
 &  &  &  & DULLAHAN~\cite{pu2024stealthy} &  & \cmark & \xmark &  & \xmark & \cmark & \xmark & \xmark &  & \fullcirc & \textendash &  &  & \fullcirc & \textendash &  & \xmark & \cmark &  & \cmark & \xmark &  & \xmark & \cmark &  & \begin{tabular}[c]{@{}l@{}}Directly injecting \\ trigger embedding into\\  the server network\end{tabular} \\
 &  &  &  & SLADV\cite{fan2023robustness} &  & \cmark & \xmark &  & \xmark & \cmark & \xmark & \xmark &  & \emptycirc & \emptycirc &  &  & \emptycirc & \emptycirc &  & \xmark & \cmark &  & \textendash & \textendash &  & \xmark & \cmark &  & \begin{tabular}[c]{@{}l@{}}Shallow model training and\\  local adversarial attack\end{tabular} \\ \hline
\multirow{16}{*}{IA} &  & \multirow{5}{*}{FIA} &  & SDAR\cite{zhu2023passive} &  & \xmark & \cmark &  & \cmark & \cmark & \xmark & \xmark &  & \halfcircL & \emptycirc &  &  & \fullcirc & \fullcirc &  & \xmark & \cmark &  & \cmark & \cmark &  & \xmark & \cmark &  & \begin{tabular}[c]{@{}l@{}}Simulator Decoding with\\  adversarial regularization\end{tabular} \\
 &  &  &  & \begin{tabular}[c]{@{}l@{}}Faeture\\Sniffer\end{tabular}\cite{luo2023feature} &  & \xmark & \cmark &  & \xmark & \cmark & \xmark & \xmark &  & \textendash & \fullcirc &  &  & \fullcirc & \textendash &  & \xmark & \cmark &  & \cmark & \xmark &  & \cmark & \xmark &  & \begin{tabular}[c]{@{}l@{}}Knowledge distillation\\  on auxiliary dataset to transfer\\  identity information\end{tabular} \\
 &  &  &  & Exact\cite{qiu2023exact} &  & \xmark & \cmark &  & \xmark & \cmark & \xmark & \xmark &  & \fullcirc & \fullcirc &  &  & \fullcirc & \emptycirc &  & \xmark & \cmark &  & \cmark & \cmark &  & \xmark & \cmark &  & \begin{tabular}[c]{@{}l@{}}Exploits activations, gradients,\\  and model parameters to\\  reconstruct private data\end{tabular} \\ \cline{3-31} 
 &  & \multirow{11}{*}{LIA} &  &Fu \textit{et al.}~\cite{fu2022label} &  & \cmark & \cmark &  & \cmark & \xmark & \cmark & \xmark &  & \textendash & \textendash &  &  & \fullcirc & \fullcirc &  & \cmark & \xmark &  & \xmark & \cmark &  & \cmark & \cmark &  & \begin{tabular}[c]{@{}l@{}}Use gradients\\  during training\end{tabular} \\
 &  &  &  & Exploit~\cite{kariyappa2021exploit} &  & \cmark & \xmark &  & \cmark & \xmark & \xmark & \xmark &  & \textendash & \textendash &  &  & \textendash & \textendash &  & \cmark & \xmark &  & \xmark & \cmark &  & \xmark & \cmark &  & \begin{tabular}[c]{@{}l@{}}Combine gradient matching \\ and regularisation term\end{tabular} \\
 &  &  &  &Liu \textit{et al.}~\cite{liu2023distance} &  & \cmark & \xmark &  & \cmark & \xmark & \xmark & \xmark &  & \textendash & \textendash &  &  & \textendash & \textendash &  & \cmark & \xmark &  & \xmark & \cmark &  & \cmark & \xmark &  & \begin{tabular}[c]{@{}l@{}}Use euclidean distance\\  and transfer learning\end{tabular} \\
 &  &  &  &Liu \textit{et al.}~\cite{liu2024similarity} &  & \textendash & \textendash &  & \cmark & \cmark & \xmark & \xmark &  & \textendash & \textendash &  &  & \textendash & \textendash &  & \cmark & \cmark &  & \xmark & \cmark &  & \cmark & \cmark &  & \begin{tabular}[c]{@{}l@{}}Cosine and euclidean \\ similarity for gradients and \\ smashed data\end{tabular} \\
 &  &  &  &Xie \textit{et al.}\cite{xie2023label} &  & \xmark & \cmark &  & \cmark & \xmark & \xmark & \xmark &  & \halfcircL & \emptycirc &  &  & \emptycirc & \fullcirc &  & \cmark & \xmark &  & \xmark & \cmark &  & \xmark & \cmark &  & Learning based approach \\
 &  &  &  & SplitAUM~\cite{zhao2024splitaum} &  & \xmark & \cmark &  & \cmark & \xmark & \xmark & \xmark &  & \emptycirc & \emptycirc &  &  & \textendash & \fullcirc &  & \cmark & \xmark &  & \xmark & \cmark &  & \cmark & \cmark &  & \begin{tabular}[c]{@{}l@{}}Using auxiliary model \\ and dummy labels\end{tabular} \\ \hline
\multirow{14}{*}{MIA/DRA} &  & \multirow{5}{*}{MIA} &  & UnSplit\cite{erdougan2022unsplit} &  & \xmark & \cmark &  & \cmark & \xmark & \cmark & \xmark &  & \fullcirc & \emptycirc &  &  & \emptycirc & \emptycirc &  & \textendash & \textendash &  & \cmark & \cmark &  & \cmark & \xmark &  & \begin{tabular}[c]{@{}l@{}}Train a functionally similar \\ clone of client's model\end{tabular} \\
 &  &  &  & SpliSPY~\cite{fufocusing} &  & \cmark & \xmark &  & \cmark & \cmark & \xmark & \xmark &  & \textendash & \textendash &  &  & \textendash & \textendash &  & \xmark & \cmark &  & \cmark & \xmark &  & \xmark & \cmark &  & use fake sample detection \\
 &  &  &  & ZETA~\cite{khowaja2024zeta} &  & \cmark & \xmark &  & \cmark & \xmark & \cmark & \xmark &  & \emptycirc & \emptycirc &  &  & \halfcircL & \textendash &  & \xmark & \cmark &  & \cmark & \xmark &  & \xmark & \cmark &  & Generative Adversarial Network \\ \cline{3-31} 
 &  & \multirow{9}{*}{DRA} &  & FORA~\cite{xu2024stealthy} &  & \xmark & \cmark &  & \cmark & \xmark & \xmark & \xmark &  & \emptycirc & \emptycirc &  &  & \fullcirc & \textendash &  & \xmark & \cmark &  & \cmark & \xmark &  & \cmark & \xmark &  & \begin{tabular}[c]{@{}l@{}}Substitute client through\\  feature level transfer learning\end{tabular} \\
 &  &  &  & UIFV~\cite{yang2024uifv} &  & \xmark & \cmark &  & \textendash & \textendash & \textendash & \textendash &  & \emptycirc & \empty &  &  & \fullcirc & \textendash &  & \textendash & \textendash &  & \cmark & \xmark &  & \xmark & \cmark &  & \begin{tabular}[c]{@{}l@{}}Construct an Inverse net to \\ extract original data information\end{tabular} \\
 &  &  &  & FSHA~\cite{pasquini2021unleashing} &  & \cmark & \xmark &  & \cmark & \xmark & \xmark & \xmark &  & \halfcircL & \empty &  &  & \fullcirc & \emptycirc &  & \xmark & \cmark &  & \cmark & \xmark &  & \xmark & \cmark &  & \begin{tabular}[c]{@{}l@{}}Hijack the learning\\  process of the model\end{tabular} \\

 &  &  &  & SIA~\cite{yu2024sia} &  & \cmark & \xmark &  & \cmark & \xmark & \xmark & \xmark &  & \fullcirc & \fullcirc &  &  & \fullcirc & \fullcirc &  & \xmark & \cmark &  & \cmark & \cmark &  & \xmark & \cmark &  & \begin{tabular}[c]{@{}l@{}}Feature concatenation\\ Gradient manipulation\end{tabular} \\ 
 &  &  &  & PCAT~\cite{gao2023pcat} &  & \xmark & \cmark &  & \cmark & \xmark & \xmark & \xmark &  & \halfcircL & \emptycirc &  &  & \fullcirc & \fullcirc &  & \xmark & \cmark &  & \cmark & \cmark &  & \cmark & \xmark &  & \begin{tabular}[c]{@{}l@{}}Construct a pseudo-client mode \\ and intermediate model to mimic \\ client functionality\end{tabular} \\
 &  &  &  & GLASS~\cite{li2024gan} &  & \xmark & \cmark &  & \cmark & \xmark & \xmark & \xmark &  & \fullcirc & \fullcirc &  &  & \halfcircL & \emptycirc &  & \xmark & \cmark &  & \cmark & \xmark &  & \cmark & \xmark &  & \begin{tabular}[c]{@{}l@{}}GAN-based Latent Space\\  Search approach\end{tabular} \\ \hline
\multicolumn{3}{l}{PA   --   Privacy Attacks} &  & & \multicolumn{4}{l}{MA --  Model Assumption}     &  &  & \multicolumn{4}{l}{Data Assumpton} &  &  &  & \multicolumn{3}{l}{MS  --  Malicious} &  &  & \multicolumn{4}{l}{GL  --  Goal} &  &  & \multicolumn{2}{l}{PS  --  Phase} \\
\multicolumn{3}{l}{PA1 --  Active Attacks} &  & & \multicolumn{4}{l}{MA1  --  Architecture}   &  &  & \multicolumn{4}{l}{DA1  --  Auxiliary Features} &  &  &  & \multicolumn{3}{l}{MS1  --  Client} &  &  & \multicolumn{4}{l}{GL1  --  Features} &  &  & \multicolumn{2}{l}{PS1  --  Inference} \\
\multicolumn{3}{l}{PA2 --   Passive Attacks} &  &  & \multicolumn{4}{l}{MA2. --  Weights}    &  &  & \multicolumn{4}{l}{DA2  --  Auxiliary Label} &  &  &  & \multicolumn{3}{l}{MS2  --  Server} &  &  & \multicolumn{4}{l}{GL2  --  Labels} &  &  & \multicolumn{2}{l}{PS2  --  Training}
\end{tabular}
}
\end{table*}

\section{Privacy-preserving Split Learning Methods}
\label{sec:defense}

The PPSL methods can be categorized into (see~\autoref{fig:PPML}): 

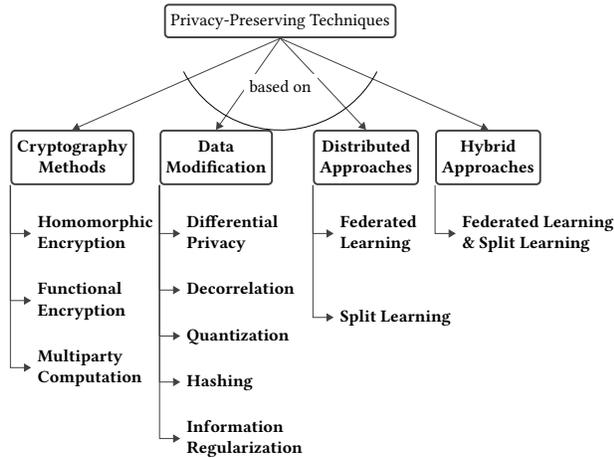
\begin{figure}[!ht]
\centering
\resizebox{0.49\textwidth}{!}{
	\begin{forest}
		for tree={
			line width=1pt,
			if={level()<2}{
				my rounded corners,
				draw=linecol,
			}{},
			edge={color=linecol, >={Triangle[]}, ->},
			if level=0{%
				l sep+=1.3cm,
				align=center,
				parent anchor=south,
				tikz={
					\path (!1.child anchor) coordinate (A) -- () coordinate (B) -- (!l.child anchor) coordinate (C) pic [draw, angle radius=20mm, every node/.append style={fill=white}, "based on"] {angle};
				},
			}{%
				if level=1{%
					parent anchor=south west,
					child anchor=north,
					tier=parting ways,
					align=center,
					font=\bfseries,
					for descendants={
						child anchor=west,
						parent anchor=west,
						anchor=west,
						align=left,
					},
				}{
					if level=2{
						shape=coordinate,
						no edge,
						grow'=0,
						calign with current edge,
						xshift=20pt,
						for descendants={
							parent anchor=south west,
							l sep+=-20pt
						},
						for children={
							edge path={
								\noexpand\path[\forestoption{edge}] (!to tier=parting ways.parent anchor) |- (.child anchor)\forestoption{edge label};
							},
							font=\bfseries,
							for descendants={
								no edge,
							},
						},
					}{},
				},
			}%
		},
		[Privacy-Preserving Techniques
		[Cryptography \\ Methods
		[
		[Homomorphic\\ Encryption ]
		[Functional \\ Encryption]
		[Multiparty \\ Computation 
		[]
		]
		]
		]
		[Data \\ Modification
		[
		[Differential \\ Privacy]
		[Decorrelation]
		[Quantization]
		[Hashing]
        [Information \\ Regularization]
		]
		]
		[Distributed\\ Approaches
		[
		[Federated \\Learning
		[]
		]
		[Split Learning
		[]
		]
		]
		]
		[Hybrid\\ Approaches
		[
		[Federated Learning \\ \& Split Learning
		[]
		]   
		]
		]
		]
	\end{forest}
	}
	\caption{Privacy-preserving Techniques}
	\label{fig:PPML}
\end{figure}
\subsection{Cryptographic based PPSL}
Well-known cryptographic techniques that enable computation over encrypted data, while preserving privacy are Homomorphic Encryption (HE)~\cite{gentry2009fully}, Multiparty Computation (MPC)~\cite{boyle2016function} and Functional Encryption (FE)~\cite{boneh2011functional}. 

\begin{center}
    \begin{tcolorbox}[width=0.49\textwidth,title={\textbf{Inherent weaknesses in SL sharing}}, colbacktitle=gray]
Many attack strategies in SL take advantage of the point where the model is split, particularly targeting the intermediate representations shared between client and server. Techniques like reconstructing the raw data from these intermediate representations have proven to be effective across different models and datasets. This suggests that privacy risks are not tied to a specific architecture or task -- they are built into the way SL operates. These vulnerabilities point to the need for model-agnostic defenses like dimensionality reduction or the use of PPT's, to better protect shared representations and gradients.
    \end{tcolorbox}
\end{center}

\noindent \textbf{HE based PPSL:} HE allows computation on encrypted data, ensuring privacy in the SL training process. This is an effective way to protect data privacy when exchanging intermediate output during the SL training process. While numerous PPML works employ HE to protect user's input~\cite{nguyen2025privacy}, there are relatively few works that combine HE with SL. Pereteanu \textit{et al.}~\cite{pereteanu2022split} propose a HE-based solution with a CKKS scheme, in a single client-server scenario. It encrypts the client's cut layer gradients that serve as input for the server layer. Khan \textit{et al.}~\cite{khan2023more} propose a protocol where the client trains his part of the model in plaintext, encrypts the intermediate output, and then sends it to the server. This ensures privacy by preventing direct access to raw data. However, this approach has limitations: during backward propagation, the server can still infer information from the gradients, posing a privacy risk. Additionally, the server-side part of the model consists of only one fully connected layer, which may limit performance. To mitigate privacy concerns during backward propagation in paper~\cite{khan2023more}, Nguyen \textit{et al.}~\cite{nguyen2023split} propose a new training protocol. Although this enhances security, the limitation of a single fully connected layer on the server-side remains. Even with HE, gradient inversion attacks can still reveal private labels with high accuracy. 
Similarly, Halil \textit{et al,}~\cite{kanpak2024cure} propose CURE, a system using HE that only encrypts the server-side model and, optionally, the input data. This selective approach, improves speed and reduces communication costs, making secure SL more practical. In addition,  PackVFL~\cite{yang2024packvfl} employs packed HE to process multiple values at once, improving overall efficiency.

Despite its strong privacy guarantees, HE-based PPSL faces challenges such as increased computational overhead. Although advances have improved, further optimizations are needed to balance complexity and privacy for real-world deployment.

\noindent \textbf{MPC based PPSL:} Multiparty Computation (MPC) provides another cryptographic method to enhance SL security against potential attacks. It enables multiple parties to jointly compute functions over their private inputs without revealing them. Khan \textit{et al.}~\cite{khan2024make, khan2025split} employ Function Secret Sharing (FSS), an MPC technique, to ensure privacy in SL. The study demonstrates that FSS eliminates privacy leakage in SL and protects against FSHA. Furthermore, this approach significantly reduces computational complexity compared to existing techniques~\cite{khan2023split}. Another key advantage of combining FSS with SL is its lower communication overhead compared to HE-based solutions, making it a more practical alternative to PPSL. Although FSS effectively eliminates privacy leakage in SL while also reducing computational complexity and communication overhead, FSS-based PPSL remains far from practical implementation. With limited research exploring its feasibility, further research is needed to address its efficiency and scalability for real-world applications.

\noindent \textbf{FE based PPSL:} FE extends the cryptographic protocol by allowing computations on encrypted data while restricting the amount of information revealed. Ma \textit{et al.}~\cite{ma2023ppsfl} propose a Privacy-preserving Split Federated Learning (PPSFL) framework that splits the global model between client and server while incorporating FE to protect data and model parameters. Their approach employs Simple-input FE (SIFE) for secure forward and backward propagation and Multi-input FE (MIFE) for privacy-preserving federated averaging. Although FE-based PPSL ensures secure model training and aggregation, its reliance on integer-based encryption introduces accuracy trade-offs. Furthermore, FE remains in an early research phase, with limited implementations.

\textbf{Discussion:} Cryptographic methods offer strong privacy in SL, especially against attacks like IA and MIA, mainly during training. However, they often come with high computation and communication costs~(\autoref{tab:comparison-ppts}), making them less practical in real-world use. For example, HE ensures strong privacy but produces large ciphertext, which is the main reason for its high overhead. Similarly FSS also adds heavy costs. While FE is more efficient, it is still an emerging area. These methods also lead to a high accuracy trade-off, due to noise or complex protocols interfering with learning.


\subsection{Data modification based PPSL}

Data modification techniques serve as a key defense mechanism in PPSL by altering intermediate outputs before they are shared with the server. PPSL based on data modification includes:

\noindent \textbf{Differential Privacy (DP):} DP~\cite{dwork2006differential} has emerged as a key defense mechanism in SL to protect sensitive information from adversarial attacks while maintaining collaborative learning. By introducing controlled noise into the intermediate output, DP aims to obscure private data while still allowing the server to perform necessary computations. Although, this approach provides privacy without the heavy computation overhead, achieving an optimal balance between privacy and accuracy remains a fundamental challenge. Abuadbba \textit{et al.}~\cite{abuadbba2020can} implemented DP by applying the Laplace differential privacy mechanism to the intermediate output. Although this method effectively reduces privacy leakage, it significantly degrades the accuracy of the model. Similarly, Titcombe \textit{et al.}~\cite{titcombe2021practical} adopted a similar approach by perturbing intermediate output before sharing them with the server. To enhance privacy protection with minimal computational cost, Mireshghallah \textit{et al.}~\cite{mireshghallah2020shredder} introduce a noise injection method that integrates noise directly into the gradient-based learning process. This technique reduces the information content of transmitted data while maintaining a reasonable balance between accuracy and privacy. On the other hand, Gawron \textit{et al.}~\cite{gawron2022feature} study the resilience of SL models enhanced with DP against FSHA. Their findings indicate that DP alone does not provide sufficient defense against such attacks, as $\mathcal{ADV}$ can still infer sensitive information despite noise perturbation. Similarly, Li \textit{et al.} propose Marvell~\cite{li2021label}, a more targeted approach by designing structured noise that minimizes label leakage against $\mathcal{ADV}$, offering an effective balance between utility and privacy.

In another study, Mao \textit{et al.}~\cite{mao2023secure} introduce a perturbation-based approach using a novel activation function $R^{3}eLU$, which randomizes responses to obscure intermediate data exchanged between the client and the server. This method effectively mitigates DRA and FSHA by controlling the amount of information leaked during training.  Building on similar principles, Qiu \textit{et al,}~\cite{qiu2023defending} propose: 
Random label Extension (RLE) and the Model-based adaptive Label Extension (MLE) method -- to address the performance degradation often caused by basic perturbation-based defenses.

To address a specific privacy concern, Yang \textit{et al.}~\cite{yang2022differentially} proposed Transcript Private Split Learning (TPSL) -- a gradient perturbation-based SL framework designed to protect label information. Using a selective noise injection mechanism, GradPerturb (gradient-perturbation scheme), TPSL ensures that noise is added only in the optimal gradient direction. Experimental results demonstrate that TPSL, particularly when combined with Laplace perturbation, achieves a good trade-off between privacy and accuracy. Building on such privacy-preserving methods, recent works have explored practical applications of SL in sensitive domains. For example, Wu \textit{et al.}~\cite{wu2023split} enable collaborative model training between terrestrial and non-terrestrial networks for smart remote sensing using SL, ensuring both data and label privacy. In another work, Sun \textit{et al.}~\cite{sun2024efficient} apply SL with DP in a satellite-based distributed learning setting using graph neural networks. Their system design addresses not only data privacy, but also tackles computational efficiency, and communication reliability, showcasing the potential of SL frameworks in complex, real-world scenarios. 

\textbf{Discussion:} DP based techniques focus on defending against DRA and MIA, under a semi-honest threat model. DP offers moderate to high privacy, depending on the noise scale ($\epsilon$), smaller $\epsilon$ means better privacy but lower accuracy. It adds medium overhead due to noise injection, especially with per-sample noise or gradient clipping, but avoid heavy cryptographic costs. Accuracy trade-offs are also medium to high, as added noise can affect model performance, especially on complex or small dataset.

\noindent \textbf{Data Decorrelation:}
The primary objective of these techniques is to limit the amount of sensitive information embedded in the intermediate output, thereby minimizing the potential of DRA by $\mathcal{ADV}$. NoPeekNN~\cite{vepakomma2020nopeek} proposed by Vepakomma \textit{et al.} integrates a distance correlation loss function into the training process. This additional loss term reduces the similarity between the input data and the intermediate output, enhancing privacy without significantly compromising the accuracy of the model.

Abuadbba \textit{et al.}~\cite{abuadbba2020can} investigated the impact of adding more convolution layers to the client-side model. Their findings indicate that increasing the depth of the client-side model slightly decreases data correlation, but at the cost of higher computational overhead. Since SL is designed to offload most computational tasks to the server, excessive client-side complexity diminishes the key advantage of SL, especially for resource-constrained devices.
Pham \textit{et al.}~\cite{pham2023binarizing}, propose that other leakage metrics, beyond distance correlation, can be integrated into the loss function. However, 
Duan \textit{et al.}~\cite{duan2022combined} noticed that relying solely on a single metric may not provide adequate privacy protection, as the intermediate output can still retain excessive information. Dougherty \textit{et al.}~\cite{dougherty2023stealthy} propose the bundle-Net architecture, which aims to protect client data from inference attack. In this approach, the client partitions the input data and shares only a partial feature representation with the server, which trains the model on these transformed data representations.
To further enhance privacy, Turina \textit{et al.}~\cite{turina2021federated} introduce a hybrid FL-SL framework with a separate loss function for the client and the server. The client-side loss prioritizes privacy by adding distance correlation, while the server-side loss focuses on optimizing model performance. 

Singh \textit{et al.}~\cite{singh2021disco}, employs selective pruning of client-side intermediate representations to remove high-risk channels before transmitting the intermediate output to the server. This method has been empirically shown to prevent DRA during inference. However, pruning requires substantial computational power, making it unsuitable for low-end client devices. Continuing along this line of research, Xiao \textit{et al.}~\cite{xiao2021mixing} propose Multiple Activations and Labels Mix (MALM), a technique that enhances privacy by mixing intermediate data across different samples and generating obfuscated labels before transmission. By reducing correlation between intermediate output and raw data, MALM provides some level of protection against DRA. However, its effectiveness is significantly diminished when performing inference on individual data samples.


\textbf{Discussion:} Data decorrelation has been used to defend against MIA and PA, mainly under the semi-honest threat model, though some works also consider malicious settings (see~\autoref{tab:comparison-ppts}). It offers medium privacy by weakening the link between raw data and shared features, but does not fully prevent leakage. Since it uses lightweight methods like suppression, its computation and communication costs remain low to medium. Some approaches apply it during training, others at inference. As it keeps key features intact, the impact on accuracy is usually small. 

\noindent \textbf{Quantization} 
is used to reduce the correlation between input data and the intermediate output, thereby enhancing privacy. In line with this, one study~\cite{zheng2023reducing} explores various compression techniques -- such as cut layer size reduction, quantization, top-k sparsification, and L1 regularization -- to improve efficiency in SL. Moreover, some of these techniques have also been applied to strengthen privacy. For example, Yu \textit{et al.}~\cite{yu2019distributed} propose a stepwise activation function that makes the activation output irreversible. The effectiveness of this method depends on the choice of stepwise parameters, balancing trade-offs between accuracy and privacy. Expanding on this idea, Pham \textit{et al.}~\cite{pham2021efficient} explore the feasibility of binarizing SL, demonstrating that binarized models can maintain nearly the same performance as the original models while significantly reducing memory usage and computational costs. The authors~\cite{pham2023binarizing} further refine this approach by introducing Binarized SL, where both the local SL model and its intermediate data are quantized. This process introduces latent noise into the intermediate output, making it harder for the server to reconstruct the original raw data.

\textbf{Discussion:} It offers a practical way to protect privacy in SL by lowering the detail of intermediate output. This helps reduce computational and communication costs, while keeping accuracy loss minimal. These methods assume a semi-honest threat model, and some are applied during training, making them efficient and lightweight options for real-world use.

\noindent \textbf{Hashing:} Another promising method of preventing DRA in SL is hashing. Qiu \textit{et al.}~\cite{qiu2024hashvfl} propose applying a sign function to the intermediate output before transmitting it to the server. This transformation makes it extremely difficult to reconstruct the original data, while still allowing effective model training. To maintain the trainability of the model under the Sign function, techniques such as batch normalization and a straight-through estimator are employed. These methods help stabilize training while maintaining model accuracy, reinforcing defense against DRAs.

\textbf{Discussion:} Hashing is a lightweight and efficient way to hide intermediate data, offering medium-level protection against DRA with low computational cost. However, it success depends on how well it is implemented -- poor design can degrade model performance. The accuracy trade-off is generally moderate.

\noindent \textbf{Information Regularization:} It limits shared information between raw data and intermediate outputs to enhance privacy. Zou \textit{et al.}~\cite{zou2023mutual} propose Mutual Information Regularization Defense (MID), which restricts the amount of information contained in the intermediate transmitted output. Their theoretical analysis demonstrates that MID effectively prevents information leakage without severely impacting model performance. In a different approach, Samragh \textit{et al.}~\cite{samragh2021unsupervised} focus on selectively discarding irrelevant information rather than directly removing sensitive attributes. Their method maximizes mutual information for utility while simultaneously reducing leakage by compressing and eliminating features that are not necessary to predict target labels.  Extending these concepts, another recent work~\cite{jiang2024training} proposes simple but effective steps to improve privacy. They use a label transformation module to hide class information, apply gradient normalization to reduce differences between classes, and add noise to obfuscate the number of classes from the $\mathcal{ADV}$. Further advancing this line of work, Alhindi \textit{et al,}~\cite{alhindi2025balancing} propose a PPSL method that integrates adversarial training with channel pruning. This approach reduces mutual information between raw inputs and intermediate output, enhancing data privacy during model training.

In line with the goal of minimizing information leakage through architectural strategies, two additional techniques further contribute to this direction~\cite{he2020attacking}. The first deactivates random neurons during inference to prevent $\mathcal{ADV}$ from reconstructing raw input data from intermediate output. Second, privacy-aware DNN partitioning suggests how to split models to lower privacy risks. While not strictly based on Mutual Information, both methods help limit the link between shared outputs and private data.

\textbf{Discussion:} It aims to improve privacy by reducing how much sensitive information is revealed during training. They offer medium computational and communication costs and have been shown to defend against MIA and LIA.

\begin{table*}[!ht]
\caption{Comparative analysis of PPSL. Privacy: High -- formal privacy guarantees, Medium -- partial defenses, Low -- vulnerable to known attacks. Computation: High -- heavy computation(GPUs) Medium: moderate overhead, Low -- minimal cost. Communication: High -- frequent exchange, Medium: noticeable overhead, Low: negligible impact}
\label{tab:comparison-ppts}
\resizebox{\textwidth}{!}
{
\begin{tabular}{lllllllllllllllllll}
\hline
Approach & PPT & Paper & \multicolumn{5}{l}{\begin{tabular}[c]{@{}l@{}}\ \ \ \ \ \ \ \ \ \ \ \ \ \ \ \ \ \ \ \ \ \ \ \ Privacy\\ \ \ \ \ \ \ \ \ \ \ \ \ \ \ \ \ \ \ \ \ \ \ \ \  Attack\end{tabular}} & \multicolumn{2}{l}{\begin{tabular}[c]{@{}l@{}}Threat\\Model\end{tabular}} & \begin{tabular}[c]{@{}l@{}}Privacy\\Level\end{tabular} & \multicolumn{2}{l}{Tra/Inf} & Comp & Comm & Acc & OSI & \begin{tabular}[c]{@{}l@{}}Cl-\\Het\end{tabular} & Limitation \\ \cline{4-8} \cline{12-13}
& & & IA & MIA & DRA & PA & FSHA & SH & M & & Tra & Inf & & & & & & \\ \hline
\multirow{3}{*}{Crypto} & HE & \begin{tabular}[c]{@{}l@{}}\cite{pereteanu2022split, khan2023more, nguyen2023split}\\ \cite{kanpak2024cure, yang2024packvfl} \end{tabular} & \cmark & \cmark & \xmark & \xmark & \xmark & \cmark & \xmark & High & \cmark & \cmark& High & High & High & \cite{khan2023more, nguyen2023split} & \xmark & Heavy Cost \\
& MPC & \cite{khan2024make}& \cmark & \cmark & \xmark & \xmark & \cmark & \cmark & \xmark & High & \cmark & \xmark & Med & High & Med & \cite{khan2024make} & \xmark & Latency \\
& FE & ~\cite{ma2023ppsfl} & \cmark & \cmark &\xmark & \xmark & \xmark & \cmark & \xmark & High & \cmark & \xmark & High & High & Low & \textendash & \xmark & Scalability \\ \hline
\multirow{5}{*}{DM} & DP & \begin{tabular}[c]{@{}l@{}}\cite{abuadbba2020can, titcombe2021practical, mireshghallah2020shredder, gawron2022feature, li2021label}\\ \cite{qiu2023defending, wu2023split, sun2024efficient, yang2022differentially} \end{tabular} & \cmark & \cmark & \cmark & \xmark & \xmark & \cmark & \xmark & Med & \cmark & \cmark & Med & Med & Med& \cite{abuadbba2020can,titcombe2021practical} & \xmark & Utility Loss \\
& DC & \begin{tabular}[c]{@{}l@{}}\cite{vepakomma2020nopeek, pham2023binarizing, duan2022combined, dougherty2023stealthy}\\ \cite{turina2021federated, singh2021disco, xiao2021mixing} \end{tabular} & \cmark & \xmark & \cmark & \cmark & \xmark & \cmark & \cmark & Med & \cmark & \cmark & Med & Med & Low & \begin{tabular}[c]{@{}l@{}}\cite{vepakomma2020nopeek, singh2021disco}\\ \cite{xiao2021mixing} \end{tabular} & \xmark & \begin{tabular}[c]{@{}l@{}}Structure\\Loss\end{tabular} \\
& DQ & \cite{zheng2023reducing, yu2019distributed, pham2021efficient, pham2023binarizing} & \xmark & \xmark & \xmark & \xmark & \cmark & \cmark & \xmark & Med & \cmark & \cmark & Low & Low & Low & \cite{pham2023binarizing} & \xmark & Precision \\
& DH & \cite{qiu2024hashvfl} & \xmark & \xmark & \cmark & \xmark & \xmark & \cmark & \xmark & Med & \cmark & \cmark & Low & Low & Med & \textendash & \xmark & Info Loss \\
& MI & \begin{tabular}[c]{@{}l@{}}\cite{zou2023mutual, samragh2021unsupervised, jiang2024training} \\ \cite{alhindi2025balancing, he2020attacking} \end{tabular} & \cmark & \xmark & \cmark & \xmark & \xmark & \cmark & \xmark & Med & \cmark & \cmark & Med & Med & Low & \textendash & \xmark & Cost \\ \hline
Hybrid & FL - SL & \begin{tabular}[c]{@{}l@{}} \cite{tian2022fedbert, kortocci2022federated, li2022ressfl, khowaja2022get} \\ \cite{jia2024model, abedi2024fedsl, xia2025sfml, shen2023ringsfl} \\ \cite{kumaar2024fortifying, gajbhiye2022data, ismail2023analyzing} \end{tabular} & \xmark & \cmark & \cmark & \cmark & \xmark & \cmark & \xmark & Med & \cmark & \cmark & Med & Med & Low & \begin{tabular}[c]{@{}l@{}}\cite{tian2022fedbert, kortocci2022federated} \\ \cite{li2022ressfl, abedi2024fedsl} \end{tabular} & \cmark & \begin{tabular}[c]{@{}l@{}}Complex\\Syn\end{tabular} \\ \hline

\multicolumn{3}{l}{DP: Differential Privacy} & \multicolumn{5}{l}{OSI: Open-source Implementation} & \multicolumn{5}{l}{DQ: Data Quantization} & \multicolumn{3}{l}{M: Malicious} & \multicolumn{3}{l}{H: Hashing} \\
\multicolumn{3}{l}{MI: Mutual Information} & \multicolumn{5}{l}{Comp: Computation Cost} & \multicolumn{5}{l}{DM: Data Modification} & \multicolumn{3}{l}{Syn: Synchronisation} & \multicolumn{3}{l}{--} \\
\multicolumn{3}{l}{DC: Data Decorrelation} & \multicolumn{5}{l}{CL-Het: Client Heterogeneity} & \multicolumn{5}{l}{SH: Semi-honest} & \multicolumn{3}{l}{Tra: Training} & \multicolumn{3}{l}{Inf: Inference} \\
\multicolumn{3}{l}{ACC: Accuracy Tradeoff} & \multicolumn{5}{l}{Comm: Communication Cost} & \multicolumn{5}{l}{Crypto: Cryptographic} & \multicolumn{3}{l}{Med: Medium} & \multicolumn{3}{l}{DQ: Data Quantization} \\ \hline
\end{tabular}
}
\end{table*}
\noindent \textbf{FL-SL synergy} 
One of the earliest approach to integrate SL in FL is SplitFed framework~\cite{thapa2022splitfed}, which enhances parallel training by incorporating FL's aggregation into SL framework. This approach applies FedAvg aggregation to update both server-side and client-side models, enabling improved efficiency. 
To improve privacy, PPFSL~\cite{zheng2024ppsfl} carefully splits models and applies group normalization layers to reduce problems caused by differences in client data. FedBERT~\cite{tian2022fedbert} enables privacy-preserving pre-training of language models, and FSL-GAN~\cite{kortocci2022federated} splits GAN components to protect client data. Methods like ResSFL~\cite{li2022ressfl} and GetPrivacy~\cite{khowaja2022get} defend against MIA using attacker-aware training and gradient-based optimization.

Other works aim to balance privacy and efficiency. For example, MP-FSL~\cite{jia2024model}, employs model pruning techniques to reduce resource consumption and prevent DRA. FedSL~\cite{abedi2024fedsl} enables privacy-preserving training of models like RNNs. Similarly, SFML~\cite{xia2025sfml} applies mutual learning for traffic classification, allowing clients to share knowledge while keeping local data private. More recent solutions such as RingSFL~\cite{shen2023ringsfl} arrange client in a ring topology and improve privacy. 
Studies like~\cite{gajbhiye2022data, ismail2023analyzing} show how PA can seriously affect FL-SL performance. To address PAs, the authors~\cite{kumaar2024fortifying} propose a robust defense mechanism using statistical methods.

\textbf{Discussion:} Combining FL and SL is a promising step forward in PPML. Recent studies show that it can defend against MIA, DRA and PA. Most works assume a semi-honest threat model. The method offers moderate accuracy, with medium computational and communication costs. It also supports client heterogeneity and keeps the accuracy trade-off low, making it practical for real-world use.

\begin{center}
    \begin{tcolorbox}[width=0.49\textwidth,title={\textbf{Prominence of studied defenses}}, colbacktitle=gray]  
The literature shows that data modification-based PPSL is the most extensively studied approach to improve privacy in SL. Within this category, techniques such as data de-correlation and DP are the most widely used techniques. Compared to other methods, HE is the most used cryptographic approach in SL. Furthermore, when both privacy and efficiency are considered, a growing body of research emphasizes the integration of FL with SL, demonstrating the potential benefits of this hybrid approach.
\end{tcolorbox}
\end{center}
\section{Open Challenges and Future Directions}
\label{subsec: Takeaways}

Cryptographic methods offer strong privacy guarantees in SL, but often come with substantial computational and communication costs, making them impractical in many real-world scenarios. Furthermore, PPTs such as FSS and FE are still in the early stages of research, requiring significant advances before they can be effectively deployed in practical applications. DP-based methods effectively reduce privacy risks but can significantly degrade model accuracy if too much noise is added. Data de-correlation offers a promising way to limit information leakage in intermediate outputs; but maybe less effective during inference. Quantization reduces the quality of intermediate output to improve efficiency and privacy but come at the cost of a potential reduction in model accuracy. Likewise, hashing is a fast and simple way to hide data, but its success depends on proper implementation to balance privacy and performance.

Another critical consideration is the model configuration itself. Deep client-side models can encode detailed input information, increasing the risk of information leakage, while shallow client models can make reconstruction attacks easier. This highlights the importance of careful model design and architectural adjustments to mitigate privacy risks while maintaining accuracy.

Many existing studies assume an $\mathcal{ADV}$ with access to parts of the model or an auxiliary data. While this helps identify potential weaknesses, this often does not reflect real-world settings where attackers have limited access. Distinguishing between different $\mathcal{ADV}$ strengths is important for building practical and realistic defenses. Additionally, as can be seen in~\autoref{tab:comparison-ppts}, most work relies on semi-honest threat model, however, real-world $\mathcal{ADV}$ can be malicious -- manipulation data or colluding with others-- making many current defenses ineffective. Future work should explore stronger, more realistic threat models to improve the security of SL.

\section{Conclusion}
\label{sec: conclusion}

This SoK provides a comprehensive analysis of privacy issues in SL, systematically examining various attack vectors that exploit vulnerabilities in intermediate outputs. We explore attacks such as MIAs and inference attacks that can reconstruct raw input data from intermediate output. Furthermore, we examine data poisoning and backdoor attacks, which manipulate training data or model parameters to compromise the integrity of SL. In addition to analyzing these threats, we explore a range of defense mechanisms designed to mitigate privacy risks, assessing their effectiveness in protecting sensitive information. However, these approaches often involve trade-offs between privacy and model performance, necessitating a careful balance to ensure both privacy and model performance.

\bibliographystyle{ACM-Reference-Format}
\balance
\bibliography{sample-base}

\appendix

\end{document}